%% file: main.tex
\documentclass[journal]{IEEEtran}

\usepackage{url}
\usepackage{amsmath,amssymb,amsfonts}
\usepackage{bm}
\usepackage{algorithm}
\usepackage{cite}
\usepackage[noend]{algpseudocode}
\usepackage{graphicx}
\usepackage[table, svgnames]{xcolor}
\usepackage[usenames,dvipsnames]{pstricks}
\usepackage{epsfig}
\usepackage{amssymb}
\usepackage{subfigure}
\usepackage{amsmath}
\usepackage{lscape}
\usepackage{makecell}
\usepackage{longtable}
\usepackage{soul}
\usepackage[colorlinks = true,
  linkcolor  = mylinkcolor,
  citecolor  = mycitecolor,
  urlcolor   = Black ]{hyperref}

\usepackage{multirow}
\usepackage{ragged2e}

\usepackage{enumitem}
\setlist[itemize]{noitemsep, topsep=0pt}
\setlist[enumerate]{noitemsep, topsep=0pt}

\colorlet{mylinkcolor}{Black}
\colorlet{mycitecolor}{Red}
\colorlet{myurlcolor}{Magenta}
\newcolumntype{?}{!{\vrule width 1.5pt}}
\hypersetup{
  colorlinks = true,
  linkcolor  = mylinkcolor,
  citecolor  = mycitecolor,
  urlcolor   = Navy 
}

\graphicspath{{src/}}

\begin{document} 

\sloppy

\title{\LARGE \bf Object Handovers: a Review for Robotics}


\author{V.~Ortenzi$^{1,4,5}$,
        A.~Cosgun$^{2,4}$,
        T.~Pardi$^{3}$,
        W. P.~Chan$^{2}$,
        E.~Croft$^{2,4}$,
        and~D.~Kuli\'{c}$^{2,4}$
\thanks{$^{1}$Queensland University of Technology, Brisbane, QLD, 4001, Australia}
\thanks{$^{2}$Monash University, Clayton, VIC, 3800, Australia}
\thanks{$^{3}$University of Birmingham, Edgbaston, B15 2TT, UK}
\thanks{$^{4}$ARC Centre of Excellence for Robotic Vision, Australia}
\thanks{$^{5}$Max Planck Institute for Intelligent Systems, Stuttgart, Germany}

}

%



\IEEEtitleabstractindextext{%
\begin{abstract}
This article surveys the literature on human-robot object handovers. A handover is a collaborative joint action where an agent, the giver, gives an object to another agent, the receiver. 
The physical exchange starts when the receiver first contacts the object held by the giver and ends when the giver fully releases the object to the receiver. However, important cognitive and physical processes begin before the physical exchange, including initiating implicit agreement with respect to the location and timing of the exchange. From this perspective, we structure our review into the two main phases delimited by the aforementioned events: 1) a pre-handover phase, and 2) the physical exchange. We focus our analysis on the two actors (giver and receiver) and report the state of the art of robotic givers (robot-to-human handovers) and the robotic receivers (human-to-robot handovers). We report a comprehensive list of qualitative and quantitative metrics commonly used to assess the interaction. While focusing our review on the cognitive level (\textit{e.g.}, prediction, perception, motion planning, learning) and the physical level (\textit{e.g.}, motion, grasping, grip release) of the handover, we also discuss safety. We compare the behaviours displayed during human-to-human handovers to the state of the art of robotic assistants, and identify the major areas of improvement for robotic assistants to reach performance comparable to human interactions. Finally, we propose a minimal set of metrics that should be used in order to enable a fair comparison among the approaches.
\end{abstract}

}

\maketitle

\IEEEdisplaynontitleabstractindextext

%
\IEEEpeerreviewmaketitle

\section{Introduction}
%
%
%
%
\IEEEPARstart{R}{ecent} years have witnessed a progression towards a more direct collaboration between humans and robots. The current trend of Industry 4.0 envisions completely shared environments, where robots act on, and interact with, their surroundings and other agents such as human workers and robots \cite{Ostergaard2017, Billard2019 }, enabled by technological advances in robot hardware \cite{Ajoudani2017}. The recent COVID-19 pandemic has increased the demand for autonomous and collaborative robotics in environments such as care homes and hospitals \cite{Tavakoli2020, Yang2020b}. Accordingly, Human Robot Interaction (HRI) is featured prominently in the robotics roadmaps of Europe, Australia, Japan and the US \cite{SRA2020_SPARC, AU_Roadmap, Japan2020, US_Roadmap}.
The advantages of human-robot teams are multifaceted and include the better deployment of workers to focus on high manipulation and cognitive skill tasks, while transferring repetitive, low skill, and ergonomically unfavourable tasks to robot assistants. Effective deployment of robotic assistants can improve both the work quality and the experience of human workers.

The structured nature of traditional industrial settings has facilitated the use of robots in work cells. However, a similarly successful presence of robots is yet to occur in unstructured environments (\textit{i.e.}, in factories without work cells, in households, in hospitals). For such environments, robots need a better understanding of the tasks to perform, a robust perception system to detect and track changes in the surrounding dynamic environment and smart, adaptive action and motion planning that accounts for the changes in the environment \cite{Pandey2014}. 
\begin{figure}[t!]
\centering
\includegraphics[width=3.5in]{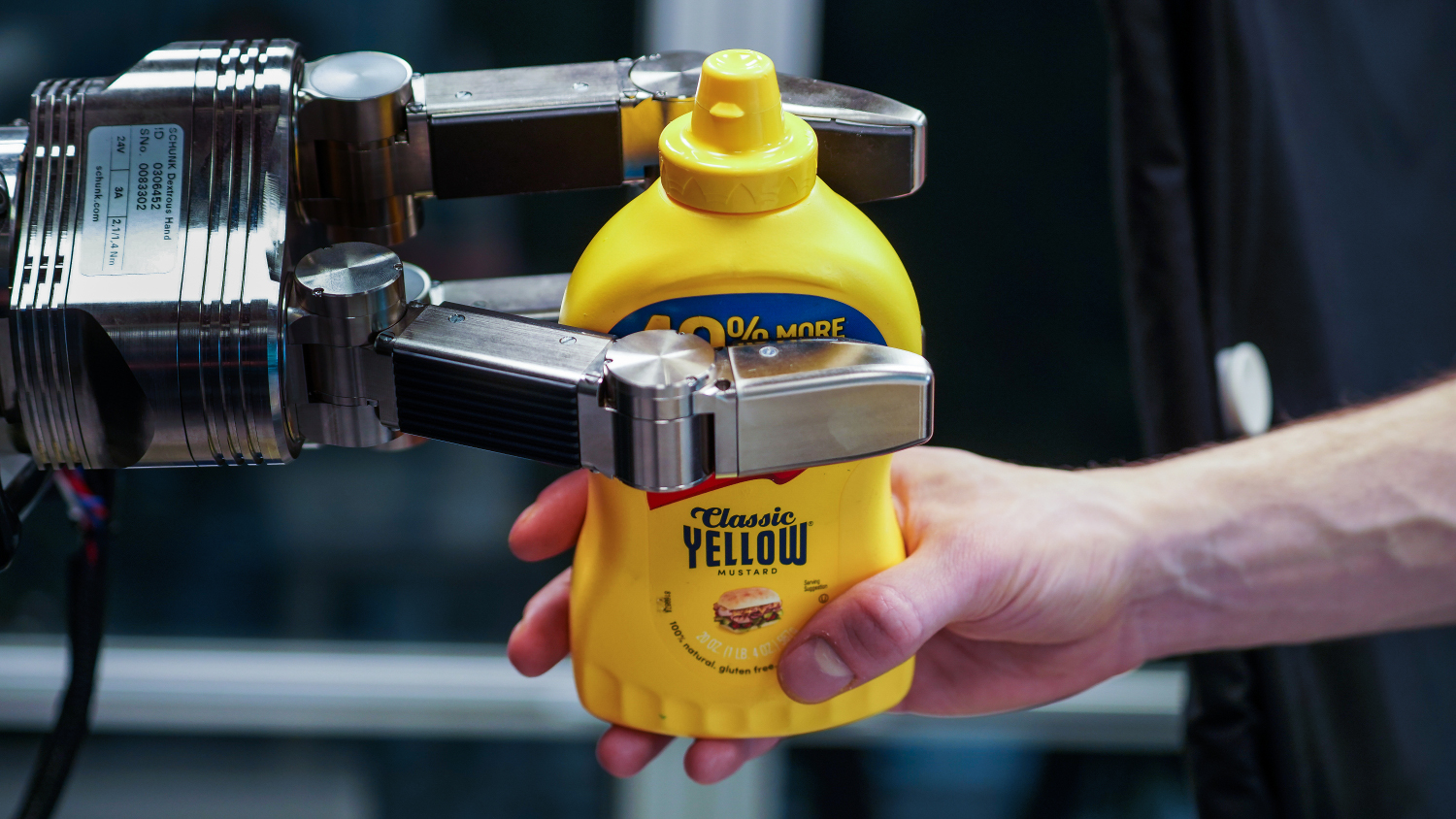}
\caption{Example of a direct handover where a robot partner passes a bottle of mustard to a human. As both hands are in contact with the object, this picture shows the physical exchange phase of a handover.
}
\vspace{-0.3cm}
\label{fig_introExample}
\end{figure}

Human-robot collaboration and human-robot interaction are frequent keywords in our research community. We refer the reader to
\cite{Thomaz2016,Ajoudani2017} for reviews on physical collaboration and to \cite{Fong2003, Freedy2007} for an overview of the cognitive aspects. 
Our community has seen an increasing focus on collaborative manipulation tasks \cite{Gu2011,Lin2018, Mielke2020}. In this context, robots must be capable of exchanging objects for successful cooperation and collaboration in manipulation tasks, as in Fig.~\ref{fig_introExample}. For example, consider an assembly task where a human operator has to assemble a complex piece of furniture and requires a tool. The robot assistant should be able to fetch and pass the tool to the human operator. Or consider a service robot handing out flyers to passersby \cite{Shi2013} or serving drinks \cite{Bohren2011}. A further example can be a mechanic asking for a tool while under a car: in this scenario, the motion range of the mechanic is extremely limited and extra care is needed to pass the tool \cite{Koene2014}.

The action of passing objects is usually referred to as an object handover. More formally, an object handover is defined as the joint action of a giver transferring an object to a receiver. This frequent collaborative action among humans requires a concerted effort of prediction, perception, action, learning, and adjustment by both parties. The implementation of a human-robot handover that is as efficient and fluent as the exchanges among humans is an open challenge for our community. In this paper, we review the state of the art of robotic object handovers. In particular, we investigate the aspects of the handover interaction that require the most effort to enable a more useful and successful collaboration with robots, particularly in unstructured environments.

We start this paper with a review of the main findings about human-human handovers in Section \ref{sec:human}. Then, in the following two sections, we refer to each of the two phases of a handover: pre-handover, and physical handover. 
In Section~\ref{sec:before} we focus on the reasoning and actions of the giver and receiver before the physical exchange of the object, analysing aspects such as communication, grasping, and motion planning and control. Section~\ref{sec:during} describes the physical exchange of the object, focusing on aspects such as grip modulation.
Section \ref{sec:safety} analyses safety in preparation and during an object exchange. Section~\ref{sec:metrics} reports a comprehensive list of quantitative and qualitative metrics that are commonly used for assessing handovers. We conclude this review with a discussion identifying open challenges and directions for future work in Section~\ref{sec:conc}. We further propose a minimal set of metrics to adopt in experimental protocols in order to enable a fair comparison among the different approaches.

\section{Human-human handover: a joint action}
\label{sec:human}
Formally, a handover is a joint action between a human giver and a human receiver. Joint actions are defined as \cite{Sebanz2006}
\begin{quote}
    any form of social interaction whereby two or more individuals coordinate their actions in space and time to bring about a change in the environment \dots successful joint action depends on the abilities (i) to share representations, (ii) to predict actions, and (iii) to integrate predicted effects of own and others’ actions.
\end{quote}
Joint actions are typically more complicated than individual actions. Social context is shown to modify the plans of actions of an agent \cite{Becchio2010}. 
While there is still much to understand and learn about how humans coordinate to meet their final goals, a number of scientific results shed some light on how humans behave during such actions. 
A minimal architecture for a joint action should include representations, processes like monitoring (feedback) and prediction (feedforward), and coordination \cite{Vesper2010}. 
Humans tend to form representations of their own goals and tasks, and potentially also of their partners' goals and tasks. Then, two processes use those representations: monitoring and prediction. Monitoring is a process to check the advancement of those tasks and goals. Such feedback can be on one's own task, on the task of the other agent \cite{Ray2018} and on the overall goal. Predicting the outcome of one's own actions and possibly, the other agent's actions, helps the coordination between the agents. Agents are interested in predicting: the what, \textit{i.e.}, the actions of the other and their goal; the when, \textit{i.e.}, the temporal coordination \cite{Capozzi2016}; and the where, \textit{i.e.}, the spatial distribution of common space \cite{Sebanz2009}. Shared representations help to predict the other's actions and achieve higher coordination, integrating the what, when and where. Coordination is also increased through joint attention (thus sharing perceptual inputs) \cite{Sebanz2006}. In particular, research has shown that there seem to be similar eye motor programs when performing and observing the same scene \cite{Flanagan2003, Rizzolatti2004}, thus reinforcing the link between perception and action. 

More recently, a Dyadic Motor Plan was proposed in \cite{Sacheli2018}.
This plan highlights the possibility that joint actions are based not only on active prediction of the actions of the partner, but also prediction of the effects of the actions of the partner, in a deeper effort of prediction. 

To summarise, during joint actions humans tend to plan their motions considering the partner's needs and representing and predicting the partner's actions and their outcomes \cite{Wolpert2000, Sacheli2018}. For this reason, scientists argue that humans form shared representations of the task to better predict each other's movements and to act accordingly \cite{Ray2011}. Efficiency and social cohesion are also listed as reasons to adopt such shared representations. 

Coordination is extremely important for the success of a joint action. There are two types of coordination \cite{Knoblich2011}: \textit{planned} and \textit{emergent}. Planned coordination emerges from the representations of the desired outcomes and one's own tasks and goals. Emergent coordination is independent of joint plans, and emerges from perception-action couplings. 
Considering these two types of coordination mechanisms, a joint action such as a handover requires the synergetic harmony of planned coordination for the final goal, and emergent coordination for the real-time aspects of the interaction.
%
%

From this perspective, an object handover is a joint action where two agents collaborate to accomplish the transition of the object from one agent, referred to as \textit{giver}, to a second agent, referred to as \textit{receiver}. While the two agents share the overall goal of the object transfer, the objectives of the two agents differ during the interaction \cite{Mason2005}. The giver aims to: most appropriately present the object to the partner; hold the object stably till the completion of the physical handover; and finally, release the object to the receiver as safely as possible. Conversely, the receiver aims to: acquire the object by grasping; stabilise the grasp on the object; and finally, following the handover, perform the task the object was required for. It is crucial to remember that, in most cases, the object is passed in order to have the receiver perform a certain task. This task might be as simple as to place the object on a table (thus imposing loose constraints on the use of the object); or it might be more complicated, such as turning a key in a keyhole or cutting a piece of paper with a pair of scissors. While these tasks are frequently actualised in our everyday life, they require an appropriate utilisation of the object, \textit{i.e.}, they impose severe constraints on the use of the object. 
The giver should consider the subsequent task that the receiver would perform with the handed over object, in order to facilitate the task of the receiver \cite{Meyer2013}.

\begin{figure*}[t!]
\centering
\includegraphics[trim= 0 25 0 25,clip, width=6.75 in]{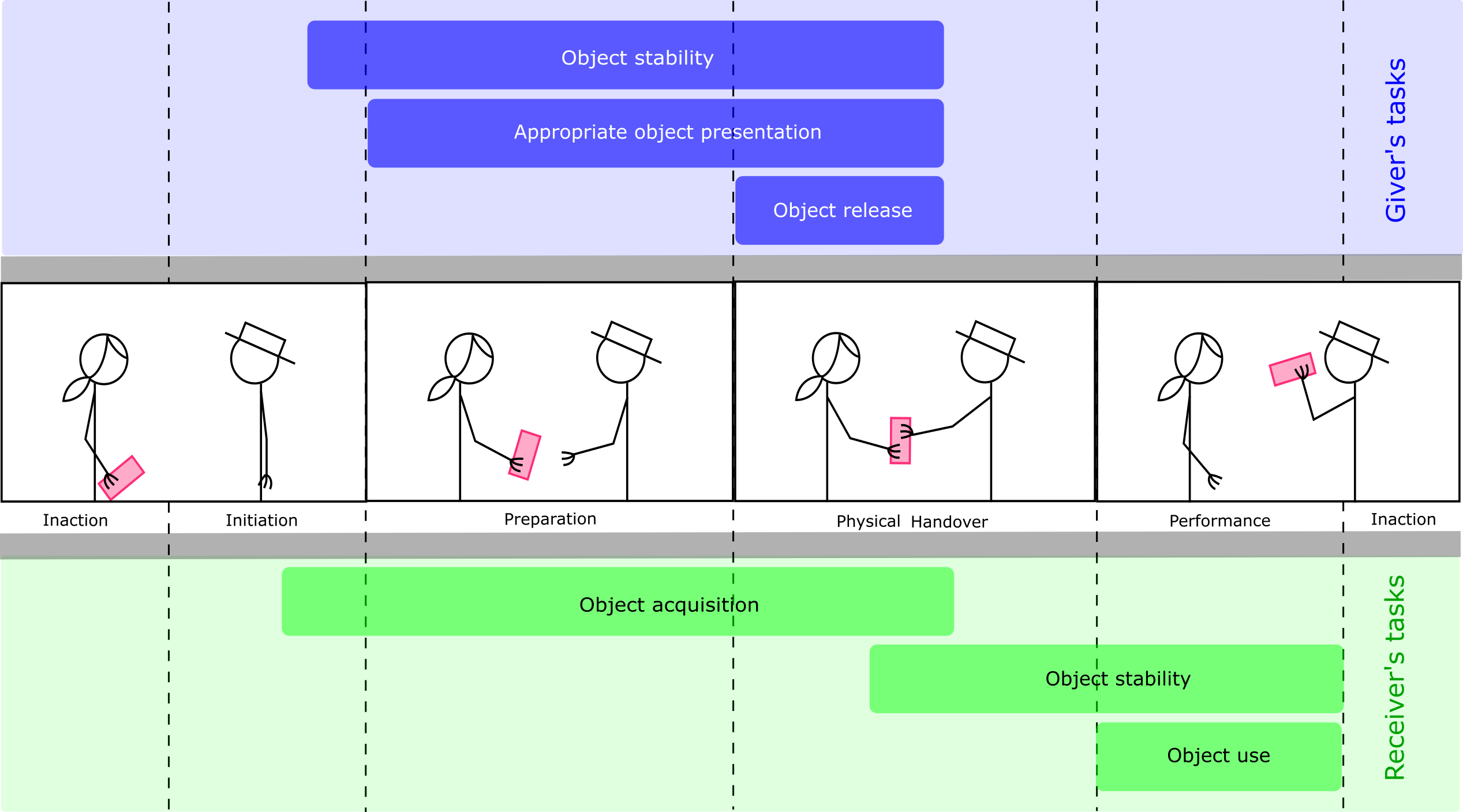}
\caption{For each phase of the handover, this figure describes giver's and receiver's tasks. }
\label{fig:activities}
\vspace{-0.5cm}
\end{figure*}

A handover can be divided into two phases \cite{Cutkosky1997, MacKenzie1994, Mason2005}. We use the tactile events, control discontinuities, and transitions that characterise any manipulation, to detail each phase \cite{Cutkosky1997}. 
Pre-handover phase includes the explicit and implicit communication between agents, as well as the grasping and transport of the object by the giver. The first contact of the receiver's hand on the object begins the physical handover. This phase comes to an end when the giver removes their hand from the object and the object is fully in the hold of the receiver. Therefore, we divide a handover into two phases: a pre-handover phase, and the physical handover phase.  During these phases, the agents display different levels of activity, with respect to their own tasks and objectives, Fig.~\ref{fig:activities}. 

Two conditions define the start and the end of a handover. A handover can be initiated by the need of an agent to obtain an object to perform a certain task (\textit{handover by object request}). This agent becomes the receiver and requests the object from the giver. The mechanic under a car asking for a tool is a typical example of this type of initiation. Another example is a cook that asks the sous chef for a kitchen tool. Alternatively, a handover can be initiated by an agent asking another to perform a certain task with an object (\textit{handover by task request}). This agent becomes the giver and gives the object to the receiver. For example, while tidying up a room, an agent can pass an object to another agent in order for the latter to place the object in a certain location; another example is a chef asking the sous chef to stir some sauce on a pan by offering the appropriate kitchen tool.

Once the exchange is initiated, the giver offers the object to the receiver. The physical exchange of the object can be \textit{direct} or \textit{indirect}. The object is passed from the hand of the giver to the hand of the receiver during a direct handover. In a number of situations, as in the example of the mechanic located under a car asking for a tool, a direct handover is also the most immediate solution to pass the requested tool. Alternatively, the object might be placed by the giver on a surface, \textit{e.g.}, on a table, during an indirect handover.
Indirect handovers allow a greater flexibility to the receiver in terms of the timing and of the grasp used to obtain the object. However, direct handovers can reduce the effort of the receiver in terms of motions required to obtain the object \cite{Choi2009}. In this paper, we focus on direct handovers because almost all the works in the robotics field belong to this kind of exchange type.

The physical phase of the handover terminates when the receiver has fully obtained the control of the object. At this stage, the receiver progresses to performing the task that initiated the handover.

The next two sections focus on action and cognition during each of the two phases of the exchange: the \textit{pre-handover}, and the \textit{physical handover} phases. In particular, we will bring attention to aspects such as  motion planning and control, prediction and communication, object grasping and offering, and modulation of grip forces. 

\section{Pre-handover phase}
\label{sec:before}
As we discussed in the previous section, a handover is initiated either by the request for an object or by the request for a task. In both circumstances, the request for an object or the request for the task must be communicated to the other agent. \textit{Communication} is a foundation for every joint action, and it can occur in various manners. Humans display a wide array of communication skills to help coordinate the \textit{what}, \textit{when} and \textit{where} of a handover. Gaze, pose and oral cues are common ways for agents to communicate during this phase. Communication does not happen only directly, as in for example voicing the intent to pass an object; but also during the action, \textit{e.g.}, in  motions or gestures during the pre-handover phase, where a giver clearly displays their intent to hand over the object. Similarly, the way an object is grasped and offered often presents cues on the intent to hand over. 

Once initiated, a handover enters the preparation phase that leads to the physical exchange of the object. 
%
%
%
%
In preparing to offer the object, the giver predicts how the receiver would perform the task the object is being passed for, and given these predictions, how the receiver would want to grasp the object. Using these predictions, the giver \textit{plans motions} to obtain (\textit{grasp}) the object if not yet grasped, or (if needed) to re-grasp the object to best prepare for the exchange, and then to offer it to the receiver. The giver relies on visual and tactile feedback to perceive and track the object as well as the state of the receiver, \textit{i.e.}, both the position and whether they are ready to receive. 
During this time, communication signals are constantly exchanged between giver and receiver. 
The giver then uses this sensed feedback to adjust their \textit{motion plans}, coupling this feedback with updated predictions of the receiver's behaviour. These updates and adaptations aided by prediction, perception and learning are used to control the motions realised to grasp the object and to offer the object to the receiver. 

The receiver shows lower activity in this phase.
However, the receiver’s actions and communication are perceived by the giver, therefore influencing the giver’s actions. Attention and state of preparedness of the receiver are important as they communicate the readiness to receive. Similar to the giver, the receiver also predicts the behaviour of the giver, and forms a plan of action. The receiver may move their hand towards the predicted handover location in anticipation of the handover. The receiver's plan and actions are updated using sensor feedback such as vision, touching and hearing. The receiver's plan and actions are also dependent on the subsequent task that the receiver would perform with the object. At the end of the pre-handover phase, the receiver has reached for and made contact with the object.

\subsection{Communication}
\label{subsec:comm}
Communication is crucial in any joint action.
Signalling strategies (\textit{i.e.}, communication) aid coordination by improving the partner's prediction of one's actions (thus minimising uncertainty) \cite{Pezzulo2013}. 
In particular, communication is used to initiate the action, \textit{i.e.}, to show the intent to start with the action; and then to coordinate the action once it has started \cite{Vesper2010}. Humans are extremely skilful in communicating their intent (the what, \textit{i.e.}, the action to perform and the object to pass) and expressing cues about the when and where of a handover \cite{Strabala2013}. Communication is so important that a handover can be thought of as a physical process (approach, reach, transfer) and a cognitive process to establish what, when and where to pass \cite{Strabala2013}. These findings indicate that robots also require such communication skills and adaptation capacity in order to match human performance during interaction with a human partner. 

Speech\footnote{Interestingly, there is evidence for the embodiment of language, \textit{i.e.}, that the motor system is activated during the comprehension of the language \cite{Jirak2010}. Moreover, there is further evidence of the involvement of the motor system in processing action words such as ``kick", ``pick". However, it is not clear yet if this activation is due to the real processing of the action words or rather it is a by-product of imagining the action \cite{Willems2007}.} can be used to express the intent to hand over an object as well as to coordinate the actions during the exchange. Speech can be used to initiate the action by either one or both of the agents, and language use could be considered as a form of joint action \textit{per se} \cite{Clark1996}. 
However, the use of speech can also degrade coordination during a joint action when the partners' attention is divided between multiple modalities of sensory communication (visual and auditory in \cite{Masumoto2014}). Similarly to human-human conversation, in HRI a robot and a human could have a dialogue to decide their roles during an interaction, and then to coordinate actions \cite{Foster2006}. 

Gaze is also a very powerful tool for communicating the intent to act and for coordinating the action. Gaze is the ensemble of eyes, head, and body orientation that reacts to the joint action \cite{Mutlu2009}. Human gaze supports the planning of actions of object manipulation, spotting positions (contact points) to which to direct a grasp \cite{Johansson2001}. Furthermore, there seems to be a link between action perception and execution. In other words, humans are able to read other people's action intentions by observing their gaze \cite{Castiello2003}. Analysing implementations of gaze in human-robot handovers, it is not surprising that during a handover, the use of gaze by a robot positively impacts the interaction, resulting in faster object reaching and a more natural perception of the interaction by the human receivers \cite{Staudte2008, Moon2014, Gharbi2015}. Similarly, gaze can have an effect on cooperation also in terms of faster human response times \cite{Boucher2012}. Interestingly, a deliberate delay in releasing the object by the robot results in an increase of attention to the robot's head, and also an increase of the compliance with the robot's suggestions (actualised with the robot's head motions) \cite{Admoni2014}. A closely related concept is turn-taking, which helps humans communicate their understanding and control of the turn structure to a conversation partner by using speech, eye gaze, and body language. Turn-taking has been explored in human-robot interaction \cite{chao2010turn}; it can be beneficial for handovers in both directions, robot-to-human and human-to-robot.

In addition to speech and gaze, humans use a number of other ways such as body stance and position, arm pose, and gestures (with arm and/or hand) to communicate their intent to hand over an object and when/where the handover will take place.
%
The presentation of an object, such as an extended arm and offering the object such that the free part is towards the receiver and tilting the object towards the receiver, are configurations that convey intent to pass an object \cite{cosgun2015did, Cakmak2011b}. Cakmak \textit{et al.} \cite{Cakmak2011b} claim that such anticipation in the behaviour of the agents makes the interactions more fluent.

In the robotics community, some aspects of such communication methods have been investigated. An analysis of kinematic features could lead to an automatic detection of the intent to hand over an object, for example using machine learning classifiers \cite{Pan2017}.
A learning-based approach presented in \cite{Strabala2012} posits that the orientation of a person and joint attention (on the object or on the position where the handover will happen) are important cues for physical interaction. Similarly, statistical models were used to model the physical aspects of a handover, and endowed with a higher-level cognitive layer that uses non-verbal cues (head orientation) to better understand the intent of a human receiver to grasp an object \cite{Grigore2013}.

Alterations of more common movements and arm trajectories can also be used by humans to communicate during joint actions.
Trajectories of motion can be altered in order to communicate to one's partner \cite{Pezzulo2013}. Taking this to the extreme, some movements can be coordinated in order to mislead one's opponent in a competitive joint action, \textit{e.g.}, a footballer's feint move \cite{Tomeo2012}. Similarly, robots can devise deceptive motions too \cite{Dragan2014}.
Moreover, the initial pose of a robot receiver can inform the human giver about the geometry of a handover \cite{Pan2018}. 

Recently, projection methods have been used for communicating the robot's intent to humans. Visualising the object pose and robot's intended grasp pose for human-to-robot handovers is shown to substantially improve the subjective
experience of the users \cite{newbury2021visualizing}.

\subsection{Grasp Planning}
\label{subsec:grasping}
We have previously considered that during a handover, the giver plans their motions considering the task of the receiver. In particular, the giver considers how to grasp the object so as to offer it to the receiver in the best way possible, \textit{e.g.}, whenever possible, to minimise object manipulation by the receiver before using the object for its intended use \cite{Gonzalez2011}. This is an example of second-order planning for object manipulation, which is defined as:
\begin{quote}
    ... altering one's object manipulation
    behaviour not just on the basis of immediate task demands but also on the basis of the next
    task to be performed \cite{Rosenbaum2012}.
\end{quote}
If the planning takes into account more than two steps, then it is termed higher-order planning. In the case of a handover, the grasp of the giver could also account for the task to be performed by the receiver \cite{Meyer2013}. 
In effect, the grasp of the giver influences the grasp of the receiver, as the latter can only grasp the object on the unencumbered portion of the object. The grasp choice of the giver can influence whether a receiver can directly use the object for their task or must re-manipulate the object to be able to use it. 

The grasping adaptation performed by the human giver is in line with theories that consider grasping an inherently task-oriented or purposive action in humans \cite{Napier1956, Feix2014corrtask, Feix2014ObjCar}, that involves both sensory and motor control systems \cite{Johansson1992, RandallFlanagan2006}. A human study shows that when participants took hold of a vertical cylinder to move it to a new position, grasp heights on the cylinder were inversely related to the height of the target position \cite{Cohen2004}, which is a clear example of adaptation of the grasp to the task. There is further evidence that the reaching movement of the arm and the grasping movement of the fingers may also be influenced by the grasper’s goal \cite{Ansuini2006, Ansuini2008, Becchio2008, Lukos2007}. From this perspective, it is not surprising that the intention to cooperate influences the grasp choice during an interaction like an object handover. As already established, givers do reason about how to grasp the object and where to place their hand on the object. Givers consider which area of the object relate to the receiver's subsequent task and adapt their grasp strategy accordingly. Indeed, when the task of the receiver has fewer performance constraints, \textit{i.e.}, when the task of the receiver is as simple as placing the object on a table, there are less stringent constraints to perform the task and thus the exchange of the object can be more relaxed \cite{Cini2019}. However, when the task of the receiver requires the use of the object in a very specific way (\textit{i.e.}, cutting a sheet of paper with a pair of scissors), then the grasp of the giver usually accounts for the constraints of the task of the receiver \cite{Cini2019}. Similarly to the considerations about the grasp of the giver, different tasks and objects elicit different levels of constraints on the grasp that the receiver has to use. 

Humans display a wide range of grasps, and several taxonomies have been proposed to categorise human grasps based on specific aspects such as hand shape on the object, contact points, and pressure \cite{Cutkosky1989, Gonzalez2014, Feix2016}. 
Humans choose their grasp considering many factors \cite{Landsmeer1962,Kamakura1980, Iberall1997, Cutkosky1989}: object constraints (\textit{e.g.}, shape, size, and function), gripper constraints (\textit{e.g.}, the human hand or gripper kinematics and the hand or gripper size relative to the object to be grasped), habits of the grasper (\textit{e.g.}, experience and social convention), and environmental factors (\textit{e.g.}, the initial position of the object and environmental constraints \cite{Puhlmann2016}). 

For all these reasons, factors such as object shape, object function and safety are important to consider when planning a grasp for a human-robot handover \cite{Kim2004}. In a human user study, it was shown that when participants are handing over objects to each other, they tend to orient some objects differently when they were explicitly asked to consider the presentation that is most convenient to the receiver \cite{Chan2015}. Similarly, object constraints and the receiver's task are highlighted to be key factors in the choice of grasp by the giver \cite{Cini2019}. In particular, grasp type and grasp location change to facilitate the grasp of the receiver on the object. Similar reasoning was already adopted for robot to human handovers in \cite{Aleotti2012, Aleotti2014}. However, the robotic giver acted knowing a priori the `appropriate' parts of the objects and the human receiver did not have to perform any subsequent task with the objects. Learning by demonstration was proposed by the same authors as a possible method to further explore the semantic segmentation of objects for grasping \cite{Aleotti2011}. Similar to this work, learning handover grasp configurations through observation of human behaviour has been shown to be a viable solution \cite{Chan2015b}. Using the concept of affordance axis, a method has been proposed for selecting good handover observation sets to learn grasp configurations \cite{Chan2019}; however, while this works well with objects with one main grasp configuration, it is a more challenging problem when the object can be presented in multiple orientations, as the robot needs to see a larger set of possible configurations and then decide which is best in a given situation.

While a successful robot grasp is usually characterised by stability \cite{Bicchi2000} and/or speed \cite{mahler2018guest}, one aspect of robotic grasping that is often overlooked is the task to perform \cite{Bohg2014} and its requirements in terms of force and mobility \cite{Ortenzi2019}. Findings in \cite{Ortenzi2020} suggest that a grasping strategy by a robot that accounts for the subsequent task of the human receiver improves substantially the performance of the human receiver in executing the following task, reducing the time to complete the task  by eliminating post-handover re-adjustments of the object. Moreover, human perceptions of the interaction improve especially when the constraints induced by the object's functional parts become more restrictive.

A planner for interactive manipulation tasks between robots could potentially account for both the grasp of the robotic giver and the grasp of the robotic receiver, thus enabling both robots to grasp successfully \cite{Lopez2006}. This approach is hardly extendable to human-robot handovers, as the human behaviour is more difficult to model with certainty. To overcome this problem, one option is to probabilistically model the behaviour of the human receiver, accounting for the ergonomic cost of the receiver, and thus influencing the grasp of the receiver \cite{Bestick2017}.

Finally, the functionality of the object being handed over is an important consideration \cite{Gibson1979, Chemero2003,Borghi2004, Schmidt2007, Thill2013, Jamone2016, Osiurak2017}. Gibson~\cite{Gibson1979} coined the term ``affordances" to define the possibilities for action offered by objects and their environment. Norman~\cite{Norman1988} added a perceptual dimension to the concept of affordance, associating it not only to the agent's capabilities, but also to their tasks to perform. However, a clear functional part of an object, such as a handle of a screwdriver, can elicit different behaviours in single-agent scenarios and cooperative tasks \cite{Sartori2011, Withagen2012, Cini2019}. 
For example, a single agent having to tighten a screw will grasp a screwdriver from the handle, whereas a giver wanting to hand over the screwdriver, should grasp it from the metal rod, thus offering the handle to the receiver. While this adaptation is natural to humans (having developed it through understanding and the repetitive use of the object), such understanding is still to be achieved in robots. A concerted effort in perception and action \cite{Bohg2017} is needed in order to endow robots with such capabilities. Learning from human demonstration and learning about physical properties of the objects that afford specific actions seem very promising approaches
\cite{Montesano2008, Nguyen2016, Detry2017, Kokic2017, Cavalli2019, Kokic2020}.
An optimisation-based approach over affordances, task to perform and mobility constraints of the human receiver is presented in \cite{Ardon2020}.

\subsection{Perception}
A big challenge in human-robot handovers is a reliable perception of the object, the hand (self and partner) and the partner's full-body motion. In this phase, vision is commonly used as the main perception channel. 
Some approaches try to track object and hand to plan for the grasp \cite{Hasson2019, Zimmermann2019, Hampali2019}, leveraging large datasets for training and physical relationships between hand and object. While grasping, the hand and objects can become severely occluded, thus harder to track with vision sensors. 
Alternatively, this problem can be addressed as a grasp classification problem \cite{Yang2020}, in which common human grasps for the task of human-robot handover are divided into categories such as ``waiting" or ``lifting", inspired by the human grasp taxonomy \cite{Feix2016}. The grasp class information can then be used by a planner to devise the most appropriate approach and grasp strategy for the robot receiver. However, the classification of grasps suffers the drawback of detecting only a relatively small subset of grasps, thus failing to detect the richness of behaviours displayed by humans. The human body can also be tracked in addition to the object and the human hand in order to improve safety \cite{Rosenberger2020}. A real-time implementation of grasp planning and re-planning for H2R handovers based on vision can be found in \cite{Yang2020reactive}. 

While the perception of the human partner's hand and body is critical real-time feedback, there have been efforts also in predicting the human partner's motion.
Dynamic Movement Primitives (DMPs) \cite{Ijspeert2002, Schaal2003} have been used successfully to predict human motion (point attractor and time scale, which mean handover location and time), coupled with an Extended Kalman Filter \cite{Widmann2018}. Real-time estimation of human motion can also leverage the concept of minimum jerk trajectories \cite{Maeda2001}. The minimum jerk model can be used in conjunction with regressors to predict when and where a human giver will transfer an object \cite{Li2015}. The minimum jerk model is used with a Semi-Adaptable Neural Network to predict human arm motion in \cite{Landi2019b}. Gaussian Processes can also be used for proactively estimating human motion for handovers \cite{Lang17}. Luo \textit{et al.} \cite{Luo2018} propose a 2-layer framework using Gaussian Mixture Models and Gaussian Mixture Regressor to represent and predict human reaching motions. 

\subsection{Handover Location}
The handover must occur in a location that is reachable by both agents. Thereby, an aspect that deserves thorough analysis is the handover location. Human-human handovers have been shown to occur roughly midway between giver and receiver \cite{Hansen2016}. Thus, the interpersonal distance between the agents has a fundamental influence on the location of the handover, and on the height of the point of exchange \cite{nemlekar2019object}. Conversely, the object mass seems not to affect the location of the exchange, but rather the duration of the exchange. 

Leveraging on this notion for HRI, a task-specific interaction workspace can be built as the intersection of the spaces that can be accessed by robot and human \cite{Vahrenkamp2016}. Information such as the effort needed by the human to reach a certain location can be used in an on-line manner to shape the interaction workspace, in order to plan the robot's movements. Similarly, handover locations can account for biomechanical properties of the human receiver, such as height, weight, strength and range of motion \cite{Suay2015}. These considerations of the biomechanical properties of the human partners are especially critical when there are environmental or task constraints to limit the motion of the human (like in the case of the mechanic under the car) and when the human is motor-impaired. Furthermore, optimising the robot's motions over safety, acceptability and task constraints could help improve the posture of the human receiver \cite{Busch2017}, thus decreasing the chances of musculoskeletal disorders and discomfort \cite{Punnett2004}. The human mobility could also be accounted for while planning, to devise different paths for the robot to the handover location \cite{Mainprice2012, Waldhart2015}. Incorporating models of the kinematics and the dynamics of the body of the human receiver can effectively devise handover locations that are more acceptable to the human partner \cite{Parastegari2017}. Finally, the human arm manipulability could also be embedded in an optimisation framework to reduce muscular strain \cite{Peternel2017, Peternel2019}. 

\subsection{Motion Planning and Control}
\label{subsec:motion}
During a joint action, the movements of the agents simultaneously actualise the physical joint action and signal important information for the coordination. 
Movements during human-human handovers are generally smooth rather than being separate and successive phases \cite{Basili2009}. For example, receivers usually start the reaching movement toward the givers while the giver reaches out for the receiver (in a concurrent motion), as implemented in \cite{Yamane2013, Medina2016}.
As such, the dominant aspects of successful movements in the context of a joint action like a human-robot handover are: legibility, predictability, safety, robustness, reactivity, and context awareness. We will cover safety specifically in Sect. \ref{sec:safety}.

\subsubsection{Legibility and Predictability} Legibility and predictability relate to how easy it is for one agent to understand and predict the other agent's movements. Albeit similar, legibility and predictability are not synonyms \cite{Dragan2013}. Using a psychological interpretation of actions, legibility is a characteristic of motion that enables an observer to infer the goal (action-to-goal). On the other hand, predictability is a characteristic of motion that matches what an observer expects given the knowledge of the goal (goal-to-action). By this definition, motions of collaborative robots must be legible, thus allowing the partner to quickly and reliably predict the goal of the actions of the robot. Interestingly, humans prefer robot configurations that are more natural or human-like as they are more readable \cite{Cakmak2011a}. Inverse kinematics algorithms mapping Cartesian motions to the robot's joint space can also aim at devising overall movements for the robot that are legible to the human partner \cite{Aristidou2011, Ortenzi2019b}. 

\subsubsection{Robustness, Reactivity and Context Awareness} The robot's motions should be flexible to accommodate changes in the environment, and to accommodate behaviours of different partners, \cite{Martinson2017, Quispe2017}. To this end, principles such as robustness, reactivity, and context awareness should guide the design of human-robot interaction systems \cite{Giuliani2010}. From this perspective, a fully pre-planned motion falls short of general adaptability. In other words, a fully deterministic approach to planning is only possible if the environment is fully known, as in the case of robot-to-robot handovers \cite{Quispe2014}. Instead, a mixture of planned motions and control over sensory feedback aids to modify the motions and adapt to the partner. A switching planning mechanism that mixes global and local planning can help to overcome the drawbacks of fixed planned motions \cite{Marturi2019}. Fast responsiveness of the robot giver is particularly important as it increases the positive impression of the interaction \cite{Koene2014}. Interestingly, a human study suggests that the speed of the interaction might be more important than the spatial accuracy of the robot for the subjective experience of a human receiver \cite{Koene2014b}. When the robot acts as a receiver, adaptive reaching displays better performance compared to a fully pre-planned reaching motion in terms of predictability and aggressiveness, \cite{Micelli2011}. Humans adapt their actions to account for the workload of their partner \cite{Huang2015}. Similarly, a robot should be aware of the task status \cite{hawkins2014anticipating}. For example, a more proactive robot giver could increase the speed of the handovers, negatively impacting the user experience. On the contrary, coordinating a reactive robot could be perceived as a better user experience, even if the performance deteriorates \cite{Huang2015}. A lower speed motion also  decreases the stress induced on the human receiver \cite{Fujita2010}.
An attempt at combining trajectories planned in the Cartesian space (emulating human movements) and joint limits was presented in \cite{Rasch2019}.

While pure planning usually devises a feedforward trajectory to follow, control architectures provide the means to use sensorial feedback and change the behaviours of the robot. Impedance control and admittance control are two common strategies to use in physical human-robot interaction 
\cite{Ott2015, Ortenzi2017, Ferraguti2019}.
Variants of classical approaches include using redundancy and null space 
\cite{Dietrich2015, Ficuciello2016}, 
modelling the interactive forces 
\cite{Magrini2015} 
and parameter adaptation 
\cite{Landi2017b}.
Early work on control proposes to use fuzzy logic on three aspects: relevance, confidence and effect \cite{Agah1997}. Human-human handovers show a smooth and fluid continuum of motion. For this reason, rather than switching control paradigms between handover phases, a phaseless controller (no distinction between reaching, passing and retracting) could be based on insights about the human behaviour, \textit{e.g.}, existence of motion during the passing and existence of coupling between the movements of the giver and those of the receiver \cite{Medina2016}. However, one specific implementation of such a controller in \cite{Medina2016} assumes that the object mass is known, in order to best modulate the grip forces. Alternatively, a controller could use high-level desired behaviours (such as proactivity or timings) using Signal Temporal Logic, as in \cite{Kshirsagar2019}.

Dynamic Movement Primitives (DMPs) \cite{Ijspeert2002, Schaal2003} represent an alternative to both pure feedforward and pure feedback control during an interaction. To specifically target a handover, the feedforward part can be weighted more at the start of the motion (shape-attraction), and subsequently the feedback (goal-attraction) can be weighted more as the interaction nears the physical exchange of the object \cite{Prada2014}. 
In order to generate a wider range of behaviours during interactions, Interaction Primitives (IPs) build on the framework of DMPs and maintain a distribution over their parameters \cite{BenAMor2014}. Probabilistic motion primitives \cite{GomezGonzalez2020} are shown to allow a robot to recognise human intent (task) and at the same time, generate commands for a robot according to the observed human motions, achieving coordination \cite{Maeda2016}. In this way, planning is replaced by inference on the probabilistic model. Learning from human feedback might also improve the adaptability of handovers.  For example, in a contextual policy search, a robot could learn a reward function from human preference feedback \cite{Kupcsik16}. Alternatively, GMMs and mirroring are proposed in \cite{Sidiropoulos2019}.

\section{Physical handover phase}
\label{sec:during}
This phase encompasses the physical interaction between giver and receiver and the object transfer. During this phase, both players are physically and cognitively engaged. 
Entering this phase, the giver possesses the object thus controls its stability. After the occurrence of the physical contact, the giver can couple vision and force feedback to understand to which extent the receiver has grasped the object. At this point, the giver starts releasing the object in order to allow the full transition of the object to the receiver. The timing must be coordinated as an early release can cause the object to fall; and a late release can cause higher interaction forces \cite{Chan2012}.

The receiver approaches this phase by planning a grasp on the object given the visual feedback from the actions of the giver. Given the presentation of the object, the receiver then acts and places the hand on the object to maximise the stability of the grasp and also in the most appropriate way to be able to perform their task afterwards. The transition ends when the giver entirely releases the object to the receiver, who then acquires the object in full. In this phase, the success of a handover is dictated by the coordination of the \textit{when} and \textit{where} of the joint action. For this reason, the most crucial aspect of this phase is the modulation of the \textit{grip force} to complete a safe transfer of the object. 
If during the pre-handover phase the main avenue of perception is vision, during the physical exchange force sensors are generally used to perceive the contacts. Other modalities of perception include tactile sensing, optical force sensing and vision.

\subsection{Grip Force Modulation}
In line with literature in neuroscience and psychology, the joint action of the physical handover is an interplay of anticipatory control and somatosensory feedback control \cite{Mason2005}. Visual feedback augments the anticipatory control in starting the release of the object, by predicting and detecting the collision created by the hand of the receiver on the object \cite{Controzzi2018}. Visual feedback is also used to adapt predictions to different speeds of the receiver’s reaching out movements. From this perspective, the speed of the grip force release seems to be correlated with the reaching velocity of the receiver (\textit{i.e.}, the faster the approach, the faster the giver releases the object) \cite{Controzzi2018}. Giver and receiver show similar strategies for controlling their grip forces with respect to the evolution of the load forces generated by the object and the exchange. All of these findings point to the fact that the giver is in charge of the safety of the object, while the receiver modulates the efficiency of the object exchange \cite{Mason2005, Chan2012}. During a human-to-robot handover forces arising during the release are different when a robot acts as receiver. In fact, the faster the retraction of the robot after grasping the object (still in the partial hold of the human giver), the larger the interaction forces. This might be explained as the giver does not have enough time to withdraw \cite{Pan2018}. 

The task of the giver is shown to resemble the evolution of a picking up task \cite{Chan2012, Chan2013} in that the giver, like the picker, typically will use excess grip force to ensure that the object does not slip or drop. Moreover, in \cite{Chan2012} a linear relationship between grip force and load force is observed, except when either actor is supporting very little of the object load \cite{Chan2012}. An analysis of these grip forces reinforces the idea that the giver is responsible for the safety of the object during the transfer, while the receiver is responsible for the timing of the transfer. A release control strategy for a robot using these insights was presented in \cite{Chan2013}. The same control strategy can also be applied to an under-actuated hand, using linear models leveraging force readings from the elbow of the robot \cite{Chan2014b}. Moreover, the feedback from a force sensor mounted on the robot's wrist can be robustly used to modulate the release of an object \cite{Konstantinova2017}. Moreover, it was shown that a proactive release improves the fluency and the subjective perception of a handover with respect to a fixed release strategy \cite{Han2019}. 

\subsection{Error Handling}
Another task for both the giver and the receiver is the handling of errors and disturbances during the handover. There might be cases where the receiver makes unwanted contact with the object and the giver should not release the object. The contact forces exerted by the receiver should then be recognised as disturbances and should be compensated to maintain a stable grasp on the object. In human-robot handovers, the tactile information from a Shadow Robot hand is used in \cite{Gomez2017} to build probabilistic models to detect these disturbances and feed them back to an effort controller. Machine learning can also be used to disambiguate among pulls, pushes, inadvertent collisions and holds performed by a human receiver on an object still in the robot's hold, as in
\cite{Davari2019}. Another threat to safety is a potential fall of the object. It has been found that human givers tend to primarily rely on vision rather than haptic sensing to detect the fall of the object during handovers \cite{Parastegari2018}. Thus, the object acceleration measured with an optical sensor at the gripper can be used as an indicator of handover failure (object dropping) \cite{Parastegari2016}.
Recently, force control and fuzzy control were similarly used \cite{Neranon2018,Neranon2019}.

\section{Safety} 
\label{sec:safety} 
Safety is a pivotal topic in human-robot interaction \cite{Haddadin2009}. In the context of a robot handover, safety is a multi-faceted concept that prioritises the physical safety of the human partner, but includes also the safe transfer of the object and the safety of the robot itself. Safety can be ensured (or achieved) through software and/or hardware \cite{Alami2006,DeSantis2008}. Research\footnote{We refer the reader to the results of the project SAPHARI, European Community's 7th Framework Programme, IP 287513, call FP7-ICT-2011-7} has led to the standard ISO/TS 15066:2016 that regulates collaborative robots and contains the norms of appropriate behaviour during physical human-robot interaction. 

The safe planning of motions while approaching a human partner is a critical aspect during a joint action. Motion planning and control can be framed to explicitly minimise safety risk during the  interaction. For instance, in \cite{Kulic2005} the robot is kept in low inertia configurations in case of unanticipated collisions; moreover, the chance of collision is reduced by distancing the robot's centre of mass from the human. Similarly, a metric of distance from the operator is used in the optimisation in \cite{Zanchettin2016}, and safety barrier functions are built around the robot links to allow collision-free planning \cite{Landi2019}.
Similarly, a safety index is used in planning augmented by human motion prediction in \cite{Wu2019}.
Motion planning should devise safe, reliable, effective and socially acceptable motions \cite{Sisbot2010, Kruse2013}. 
Frontal approach versus lateral approach by the robot towards the human receiver is discussed with some contrast in \cite{Koay2007,Walters2007}. Such considerations are further used to develop the planner in \cite{Sisbot2010}, which is composed of three components: spatial reasoning to account for the human receiver (perspective placement \cite{Urias2008}), path planning optimising over costs that account for safety, visibility and human arm comfort (human-aware manipulation planner \cite{Sisbot2007}), and trajectory control to ensure minimum-jerk motions at the end effector (soft motion trajectory planning \cite{Broquere2008}).
Humans minimise jerk in order to realise well-behaved trajectories for arm movements \cite{Flash1985}. Minimum-jerk motions by a robotic giver also result in shorter reaction time and faster adaptation for human receivers \cite{Huber2008a}. Further, to better match the human trajectories of minimum jerk, a decoupled minimum jerk trajectory could be used, using different time constants in the gravity axis z (thus decoupling the motion in the x-y plane to the motion in the z axis) \cite{Huber2009a}. 

The two paradigms R2H and H2R involve different aspects of safety. In the H2R paradigm, the robot aims to make contact and grasp only the object, avoiding any contact with the human partner. This is usually achieved leveraging vision, \textit{e.g.}, \cite{Rosenberger2020, Yang2020, Yang2020reactive}. In order to avoid any contact with the human partner, most of the approaches are over-conservative in the attempt to compensate for potential noise in the perception (\textit{e.g.}, building enlarged bounding boxes around the hands and body of the human partner, thus allowing a greater distance between robot and human).

Another aspect that is critical to safety is the grasping/pulling force exerted on the object, and its timing. An erroneous timing and/or a too high/low pulling force could generate highly unsafe behaviours such as: pulling the human partner along with the object, or allowing the object to drop \cite{Chan2012, Chan2013}. In the R2H paradigm, the robot must (i) approach the human safely (without contacting/hitting the partner) and orient the object appropriately (such as pointing the tip of a knife away, or presenting the handle of a cup of hot coffee, or not spilling any of the contents of the object, such as the coffee in the cup) \cite{Aleotti2014} and (ii) safely release the object when the human partner has grasped it \cite{Rosenberger2020}. When handing an unknown object during R2H, it may be challenging for the robot to accurately assess the danger of the object to the human receiver.

One last aspect of safety are social conventions \cite{Becchio2008, Becchio2010}. Behaviours such as handing over a knife by offering its handle are not only safer per se, but they are regarded as socially more acceptable than thrusting a blade to one's partner, which can convey an erroneous intent (not to mention the inherent risk of harming one's partner) \cite{Kim2004, Haddadin2011}. Such social conventions offer interesting insights in order to produce safer and more readable behaviours in robots \cite{Sisbot2012}.

\section{Metrics}
\label{sec:metrics}


There is a general consensus on the need for standardised measurement tools and metrics in the human-robot interaction and collaboration communities \cite{Steinfeld2006, Aly2017}. However, the spectrum of aspects to cover is so broad that finding a set of metrics and tools to adopt in every situation is very difficult. Nevertheless, such common and codified metrics would allow for an easier and fairer comparison among the proposed techniques, and would possibly help to build new frameworks. Metrics should aim to assess a handover qualitatively and quantitatively \cite{Steinfeld2006}. Along the same lines, a survey on metrics for human-robot interaction \cite{Murphy2013} reports productivity, efficiency, reliability, safety and co-activity to be the areas to assess for an interaction. Furthermore, there is a wide range of literature analysing metrics for human-robot interaction and collaboration, such as for human-robot teams \cite{Holzapfel2008,Pina2008,Burke2008, Nielsen2008} and for social and physical interaction \cite{Bartneck2008, Chaminade2008, Oosterhout2008}. 

\begin{figure}[ht!]
\centering
\includegraphics[trim= 0 0 0 0,clip, width=3.5in]{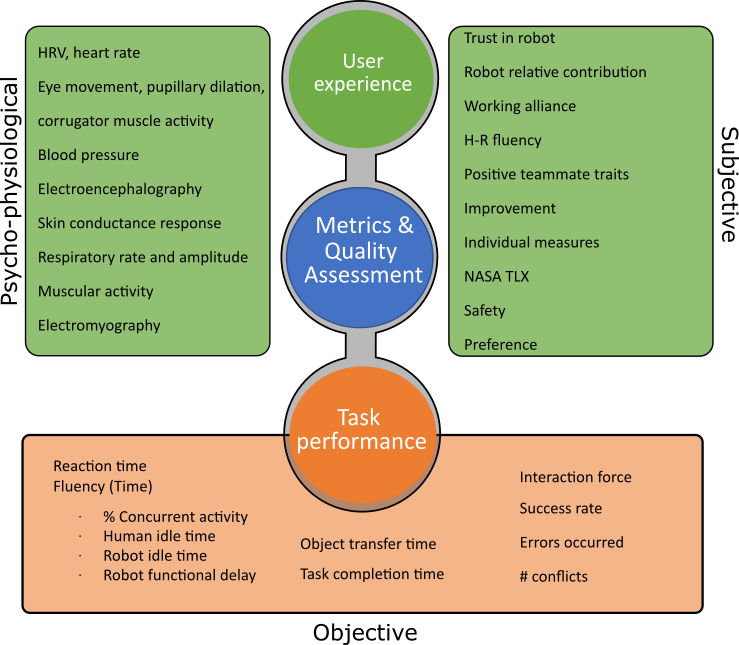} 
\caption{Metrics assess the overall performance of a handover, with measures such as timing and success rate; but also the user experience, with psycho-physiological measures and subjective measures.
}
\vspace{-0.3cm}
\label{fig_metrics}
\end{figure}

In this section we analyse three different types of metrics: 1) task performance metrics which provide a measure of success, 2) psycho-physiological metrics to measure the human partner's physiological responses, and 3) subjective metrics in the form of user questionnaires. These metrics are represented graphically in Figure \ref{fig_metrics}. We also analyse the variety of the test objects used in handover experiments.

\subsection{Task Performance Metrics}
Task performance metrics are often used in HRI experiments to evaluate success quantitatively, and the choice of such metrics is highly dependent on the task. 
The performance of a handover can be coarsely described using the success rate: number of successful handovers divided by the total number of trials. Success rate is the most popular task performance metric for human-robot handovers.
Even though the overall success rate of an implementation is important, it only reports a statistical view of the handovers rather than the quality of the interaction, and by itself, it does not explain why and how the errors have occurred. Besides, different experimental protocols make it difficult to compare the success rate metrics directly. The interaction force is another measure that has been commonly used to evaluate the success of the interaction.

Considerations of performance also include the task completion time. From this perspective, fluency is an important characteristic of an interaction such as the handover. To evaluate fluency, objective metrics should include percentage of concurrent activity, human idle time, robot idle time and robot functional delay \cite{Hoffman2007,Hoffman2007b,Unhelkar2014 ,Hoffman2019}.  These concepts are also related to task effectiveness and interaction effort \cite{Goodrich2003}. Moreover, time considerations can include the reaction time of the human, task completion time and overall handover time. Among the surveyed handover papers, time-related metrics include: waiting time of the robot and the human, total handover time and timing of different phases of the handover. Other task performance metrics used in handovers include defining and minimising a cost function related either to the trajectory or to the interaction.

\subsection{Psycho-Physiological Metrics}
Another way to gather quantitative data from user studies is to measure the physiological responses of the human partner during the interaction. In HRI, psycho-physiological measures can be used to identify and evaluate the human partner's responses to the interaction with the robot \cite{Bethel2007}.  Physiological signals such as electromyography (EMG) can be used to measure the human's motor activity during the handover.
Physiological signals can also be used to estimate the affective state of a human partner during an interaction. 
Furthermore, physiological responses can be exploited when evaluating responses to a safe planner (less anxiety and surprise, reported feeling more calm) \cite{Kulic2006}. Another example is 
Heart Rate Variability (HRV), which can be used as a quantitative index to assess mental fatigue \cite{Villani2018}. The psycho-physiological measures to assess anxiety and stress in response to the interaction include, but are not limited to: eye movement; heart rate and heart rate variability; blood pressure; electroencephalography; skin conductance response; pupillary dilation; respiratory rate and amplitude; muscular activity; corrugator muscle activity; electromyography.

\subsection{Subjective Metrics}
Subjective metrics assess aspects such as the subjective perception of the human regarding the perceived difficulty of the task, the cooperation and alliance of the robot, trust in the robot and contribution of the robot \cite{Hoffman2019}. Additional concepts that are recurrent in a qualitative evaluation of an interaction include anthropomorphism, animacy, likeability, perceived intelligence, and perceived safety \cite{Bartneck2009}. The Robotic Social Attribute Scale (RoSAS) framework proposes measuring the subjective and social perception of robots using three dimensions: warmth, competence and discomfort \cite{Carpinella2017}. Legibility, safety and physical comfort are also key criteria to consider \cite{Dehais2011}. Furthermore, ad-hoc questionnaires and the NASA-TLX \footnote{https://humansystems.arc.nasa.gov/groups/TLX/} can be utilised to provide additional instruments to assess the cognitive workload of humans. 

The most common vehicle for user studies in the reviewed papers were post-study surveys, in which the participants rated different aspects of their interaction in a Likert scale. The most commonly asked questions in the questionnaires relate to the fluency of the interaction (\textit{i.e.}, natural, legible, predictable robot motions), 
how safe 
and comfortable 
the participants felt during the interaction, whether participants were satisfied with the experience,
the ease of use of the interface, 
the competence of the robot,
 the appropriateness of the robot's timing, 
the perceived aggressiveness of the robot,
the trust in the robot, 
and whether the robot acted in a human-like manner.
In addition, for some papers the main subjective evaluation was the indication of preference and/or subjective opinions and comments from the participants.

\subsection{Test Objects}
There have been recent efforts in the grasping community to create physical benchmarks and experimental protocols in order to facilitate the replication of research results \cite{mahler2018guest,bekiroglu2019benchmarking, Negrello2020}.
Towards the same goal, object datasets have been generated for grasping, such as YCB: an object dataset \cite{Calli2017};
and DexNet: a synthetic
dataset of 6.7 million point clouds, grasps, and analytic grasp
metrics \cite{Mahler17}. 

The choice of objects used in human-robot handovers usually depends on the target application; for example it differs for industrial and domestic environments. The last column of Table~I and II shows how many test objects were used for the experiments in the correspondent papers. We found that the vast majority used only a single object class for the experiments. The most commonly used objects were cylindrical objects such as bottles, 
followed by rectangular objects such as boxes.
While some researchers opted for custom-designed objects with sensors mainly for measuring grip and load forces,
some have chosen application-specific objects such as flyers~\cite{Shi2013}.

\section{Discussion and Directions for Future Work}
\label{sec:conc}
\begin{figure*}[ht!]
\centering
\includegraphics[width=7in]{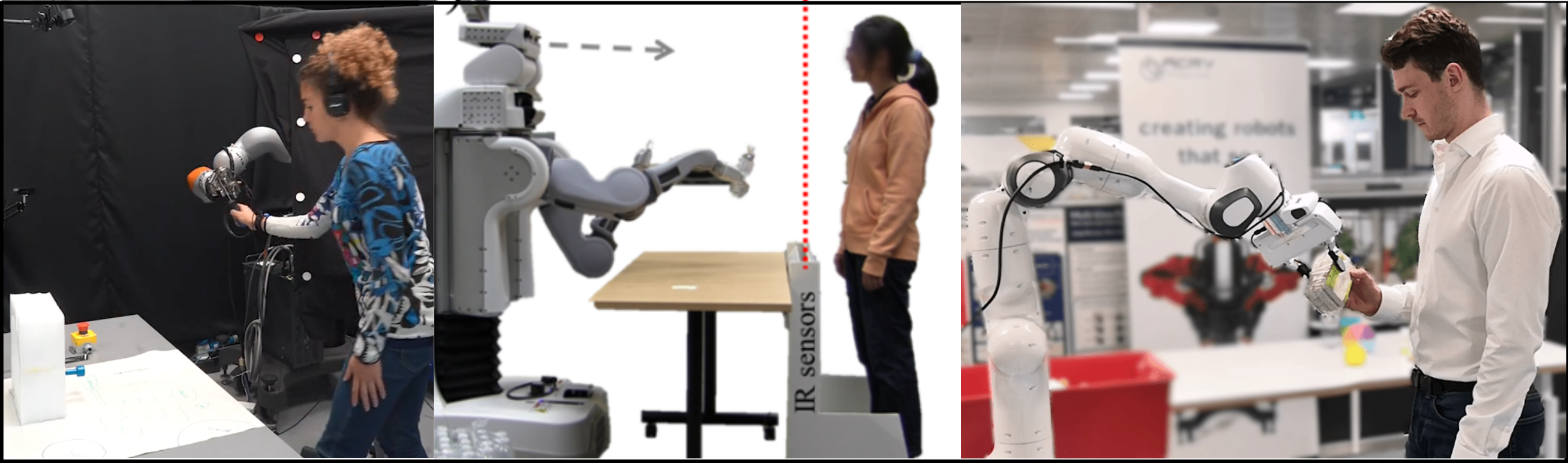}
\caption{Examples of real implementations of human-robot handovers. The first two images depict the R2H paradigm, while the rightmost image depicts the H2R paradigm. In all the three instances, a real robot is interacting with the human participant. Images are taken respectively from \cite{Ortenzi2020}, \cite{Moon2014}, \cite{Rosenberger2020}
}
\label{fig_implem}
\vspace{-14pt}
\end{figure*}

We discuss human-robot handover papers that present experiments with a real robot, as depicted on Fig.~\ref{fig_implem}. An overview of these contributions can be found in Table~I and II. For each paper we report: paradigm (R2H or H2R); what the authors investigated (communication, grasping, motion planning and control, and perception during the pre-handover phase; grip force and error handling during the physical handover); the sensors used;
whether the handover location was fixed, pre-planned 
or adapted online to the human partner; whether the experimental protocol included a post-handover task for the receiver; the metrics used to assess the task performance and the user experience; and finally the number of different objects used in the real robot experiments.

There are a few observations emerging from Table~I and II. In general, the paradigm R2H has been investigated more frequently than H2R. The handover location is usually either fixed or pre-planned; on the other hand, online adaptation is much less frequent. The physical handover phase has not been studied as frequently as the pre-handover phase. Furthermore, there is a general lack of uniformity in the experimental protocols, especially in terms of: presence of a post-handover task, metrics to assess the results, and number of test objects.

In order to bridge the gap between human-human handovers and human-robot handovers, we identify two open challenges. First, the interaction should become more fluid and fluent. In most current work, robots present predefined behaviours that force the human partner to comply. This not only decreases the perceived alliance of the robot, but makes the joint action less natural. Second, we believe that experimental protocols should be more standardised, in order to allow a fairer comparison among the proposed algorithms, methods and approaches.

\subsection{Open challenge 1: Adaptability}
\subsubsection{Adaptability and Handover Location}
Studies in neuroscience, physiology and psychology highlight that a handover is an intricate joint action that requires physical and cognitive coordination. In particular, the cognitive level of the interaction is as important as the physical level \cite{Lemaignan2014}, for a robot to be considered as a partner, and not only as a tool \cite{Breazel2004}. To match the human skills of understanding and adaptation \cite{Edsinger2007}, it is preferable that robots also display adaptation and understanding. 
In fact, human givers can control the object's position and orientation to facilitate the robotic receiver's grasping of the object \cite{Edsinger2007}.
\input{table_final}

During extended interaction, fatigue of the human worker having to accommodate the robot repeatedly and for a long time can become an issue \cite{Peternel2019}. From this perspective, robots that are able to adopt different behaviours adapting to their partner could assist their human counterpart \cite{Koene2014, Huang2015}. Moreover, different users generally interact with a robot in different ways during a handover \cite{Meyer2017}.
In other words, it is crucial to account for the feedback coming from the human partner during the interaction when controlling the robot. However, as can be seen in the Tables, most approaches focus on fixed or pre-planned handover locations, with far fewer attempts at adapting to the human partner online. In many human-robot handover scenarios, the handover location is kept fixed (the robot is either going always to the same position for the object transfer), or the handover location is pre-planned based on several criteria (including ergonomics, safety, etc.) and not updated in real-time with the perceptual feedback. This is far from ideal, as the human has to potentially adapt to the robot and could incur cognitive and physical fatigue. The ergonomics of the interaction should be accounted for, as the transfer should happen in the comfort zone of the human, \textit{i.e.}, the range of positions (and tasks to perform in) reachable (and doable) with little or no compensatory movements \cite{Kolsch2003}. In humans, an optimisation principle over a muscle stress index is shown to determine the arm motions and postures (selected over the infinite possibilities of motion) and also the perceived comfort \cite{Katayama2003}. 
We believe that while pre-planning such a location accounting for the ergonomics and the physical characteristics of the human partner is appropriate, the handover location should not be fixed or pre-planned, but adapted online to the human partner. More effort is needed in order to adjust online to changing circumstances (adapting in real-time to the needs of the human partner). 

\subsubsection{Communication}
Communication is a key factor to achieve a successful coordination during a joint action. Humans use speech, gaze, and body movements to communicate intent, and coordinate during the execution of the joint action. We observe that robots have displayed a general lack of communication skills for object handovers. Most of the effort in the literature so far has been put on the physical aspects of the interaction, focusing on motion planning and control, grasping and perception. 
On the other hand, effort in communication is less prevalent, 
(this aspect can also be noticed in the Tables, as only a minority of the papers include an element of communication in their implementation).
We believe that improving the communication cues provided to the human partner by the robot is a key factor to increase the naturalness and fluency of human-robot handovers.

\subsubsection{Grip Release}
There are also only a few papers that focus on grip release and how to handle potential falls of the object. While the literature in human studies continues to investigate how both agents modulate their grip force on the object and how the different sensory modalities (vision and tactile) come into play, most of the reviewed work has adopted a simplistic approach, \textit{i.e.}, robotic givers completely release the object whenever a pull by the receiver is detected. Conversely, robotic receivers need to modulate their pulling force as too little force could be unsafe for the object transfer and too much force could be dangerous for the human partner.
We believe that grip force modulation is a key component that needs further investigation and effort. There are many additional open research directions, such as: (i) the use of different hardware (under-actuated vs fully actuated, soft vs rigid, parallel jaw gripper vs multi-fingered hands vs suction) as in \cite{Bianchi2018}; (ii) the use of different grasping strategies (grasp type and location on the object); and (iii) the use of objects varying their size and weight. Such exploration could thus give rise to various options to modulate the grip.

\subsection{Open challenge 2: Standardised experimental protocol}
In order to enable a fair comparison among the contributions and improve human-robot handovers, a standardised experimental protocol should be developed and adopted, to generate results that are easy to interpret and easy to compare against. We propose to focus on three aspects: the post-handover task, the objects used in the experiments, and the metrics to assess the results.

\subsubsection{Role of the post-handover task}
From the robot's higher-level behaviour standpoint, there is a critical need for improvement in the integration of cognitive and physical reasoning \cite{Pandey2014} in both paradigms (H2R and R2H). In other words, robots currently lack a vision and understanding of the general goal of such an action. Such understanding is the key contributor to enabling higher-order planning \cite{Gonzalez2011, Rosenbaum2012, Meyer2013, Cini2019}.
For example, robotic grasping has achieved peaks in performance
\cite{Morrison2019,Adjigble2018,Levine2018}; however, the ultimate goal of the grasp is rarely taken into account \cite{Ortenzi2019}. As a result, robots can manage to grasp objects but seldom these grasps would allow the execution of a task with the objects. During a handover, a successful grasp should account for the interaction partner. In \cite{Sanchez-Matilla2020}, a benchmark for H2R handovers is proposed to promote a fairer comparison among algorithms, offering sub-scores for each handover phase sub-action.

Following a similar reasoning, we believe that any experimental protocol should include a task to perform by the receiver with the handed-over object, as proposed in \cite{Cini2019}. This is a critical consideration because the object exchange is normally initiated in order for the receiver to perform a task with the object. 
A complete experimental procedure should consider the capability of the receiver to use the object directly following the handover. If the receiver can grasp the object in a way that its subsequent use does not require any further re-manipulation, then the receiver can start the task straight away after the physical exchange of the object. 
Conversely, the receiver might need to re-adjust their grasp of the object in case their temporary grasp (realised during the exchange) is not an ideal grasp to correctly use the object for the specific task \cite{Ortenzi2020}.
However, this post-handover grasp adjustment could decrease the quality of the handover in objective terms (longer task performance time, higher strain when the handover happens multiple times) and in subjective terms (the giver could be perceived as a lesser partner, and the task could be perceived as more cognitively difficult). These quality evaluations are pivotal in establishing the degree of success of a handover. However, very few experimental protocols include a posterior task for the receiver. 
Even though it might be argued that a handover can be considered finished after the object transfer, we believe that such post-handover task performance is important to effectively assess the overall performance of the dyad and gauge the experience of the human partner. 

\subsubsection{Proposed Set of Metrics}
Our survey has revealed a need for standardisation in the choice of metrics and objects for real robot experiments. Most of the surveyed papers report results using task performance metrics (\textit{e.g.}, success rate and timings) and subjective metrics on the experience of the human partner (often in the form of Likert-scale post-experiment questionnaires). We believe that a minimal set of metrics should be defined in order to enable a fairer and more direct comparison among the different approaches. To this end, we propose the following combination of metrics that assess the most common aspects of a handover: 
\begin{enumerate}
    \item Task performance (objective): success rate, total handover time, receiver's task completion time.
    \item Experience of the human (subjective): fluency, trust in robot, working alliance.
\end{enumerate}
This minimal set includes metrics which are clearly defined, thus reproducible, and which are easy to measure. For these reasons, the set does not include psycho-physiological measurements as they require sensors placed on the body of the human participant, and thus are difficult to standardise and deploy in a variety of contexts. 

The experience of the human participant should be assessed administering the following questionnaire (the following set of questions includes a subset of questions from \cite{Hoffman2019}):
\begin{enumerate}
    \item Human-Robot Fluency 
    \begin{itemize}
        \item The human-robot team worked fluently together.
        \item The human-robot team's fluency improved over time.
        \item The robot contributed to the fluency of the interaction.
    \end{itemize}
    \item Trust in Robot 
    \begin{itemize}
        \item I trusted the robot to do the right thing at the right time
        \item The robot was trustworthy.
    \end{itemize}
    \item Working Alliance 
    \begin{itemize}
        \item The robot accurately perceives what my goals are.
        \item I understand what the robot's goals are.
        \item The robot and I are working towards mutually agreed upon goals.
    \end{itemize}
\end{enumerate}
All questions should be evaluated on a Likert scale.
We believe that this set of questions covers a broad set of important general aspects of the interaction, namely fluency, trust and working alliance. Furthermore, additional questions can be added to this minimal set in order to investigate additional specific aspects of a handover, such as preference between different approaches, and learning/improvement over time.

\subsubsection{Objects}
The vast majority of papers on human-robot handovers use only a single object class. This observation shows that generalisation of handovers to a variety of objects has not been the main focus of a majority of the papers until very recently \cite{Rosenberger2020, Yang2020reactive}. The most commonly used test objects have been either cylindrical objects such as bottles, or rectangular objects such as boxes. This is likely because these object shapes are easier to grasp and many everyday objects belong to these categories. 
We argue that future experiments should include a broader set of objects, as different objects generate different behaviours and can be used to address different manipulation tasks. We propose the use of objects that elicit all three grasp macro-types in \cite{Cutkosky1989}, \textit{i.e.}, power, intermediate and precision grasps. The three macro-types offer sufficient opportunities to explore different behaviours, investigating aspects such as different object offering and reception; different post-handover tasks; and handover of objects with different weights and shapes. 
Nevertheless, we acknowledge that the choice of objects might depend on the specific focus of each study. For example, studies on the reaching motion might place their focus on the motion and not on the objects, so three objects evoking the three grasp types would be enough. However, studies more focused on the objects, such as a study on object orientation in the preparation to hand over, would require a wider set of experimental objects.

Our proposition of a minimal set of metrics and of objects to use in an experimental protocol is targeted to increase the possibility of fair comparison among the approaches. A handover is a sophisticated joint action that includes many different aspects (communication, planning, grip release, etc). For this reason, there has been a general non-uniformity in protocols and metrics. We believe that our proposition of metrics and objects covers the most common aspects of a handover, thus enabling a fair comparison among approaches, while allowing for additions when the research questions call for investigation into more specific aspects.


\subsection{Consideration of paradigm}
In terms of the paradigm, R2H handovers have been more frequently investigated. We speculate that the idea of having a robot assistant that can fetch objects and give them to humans when needed, has driven the deeper investigation of the R2H paradigm. The R2H paradigm is particularly representative of the cases where the human receiver will then perform a cognitively challenging task with the object, a task that robots are not yet able to perform. However, it is our opinion that H2R handovers are worth exploring more and represent an open area of research. One of the biggest challenges in human-to-robot handovers is safety \cite{Rosenberger2020}, as the robot should be careful to not contact the human giver. For this to happen, perception systems should be able to robustly discriminate the human giver (hand and arm) from the object \cite{Yang2020, Rosenberger2020, Yang2020reactive}. Moreover, in the H2R paradigm grasp planning becomes another critical issue, as the robot will have to perform a task with the handed-over object, \textit{i.e.}, at the very least need to put the object down in a pose preferable to humans \cite{newbury2020learning}. 

\section*{Acknowledgment}
The authors want to thank the Australian Research Council (Centre of Excellence for Robotic Vision (project number CE140100016)).
Elizabeth Croft acknowledges support from Australian Research Council (project number DP200102858) and the Natural Sciences and Engineering Research Council of Canada (project number RGPIN-2017-04450).
Tommaso Pardi is supported by a doctoral bursary of the UK Nuclear Decommissioning Authority.
The authors want to thank:
Peter Corke for his invaluable feedback; Alessandro De Luca for the good advice about safety in physical human-robot interaction; Marco Controzzi for the discussion on the role of the task in experimental protocols; Maya Cakmak for recommendations to further improve this manuscript. 
\ifCLASSOPTIONcaptionsoff
  \newpage
\fi



\bibliographystyle{IEEEtran}
\raggedright
\bibliography{bib_handover}
%



%

\begin{IEEEbiography}[{\includegraphics[width=1in,height=1.25in,clip,keepaspectratio]{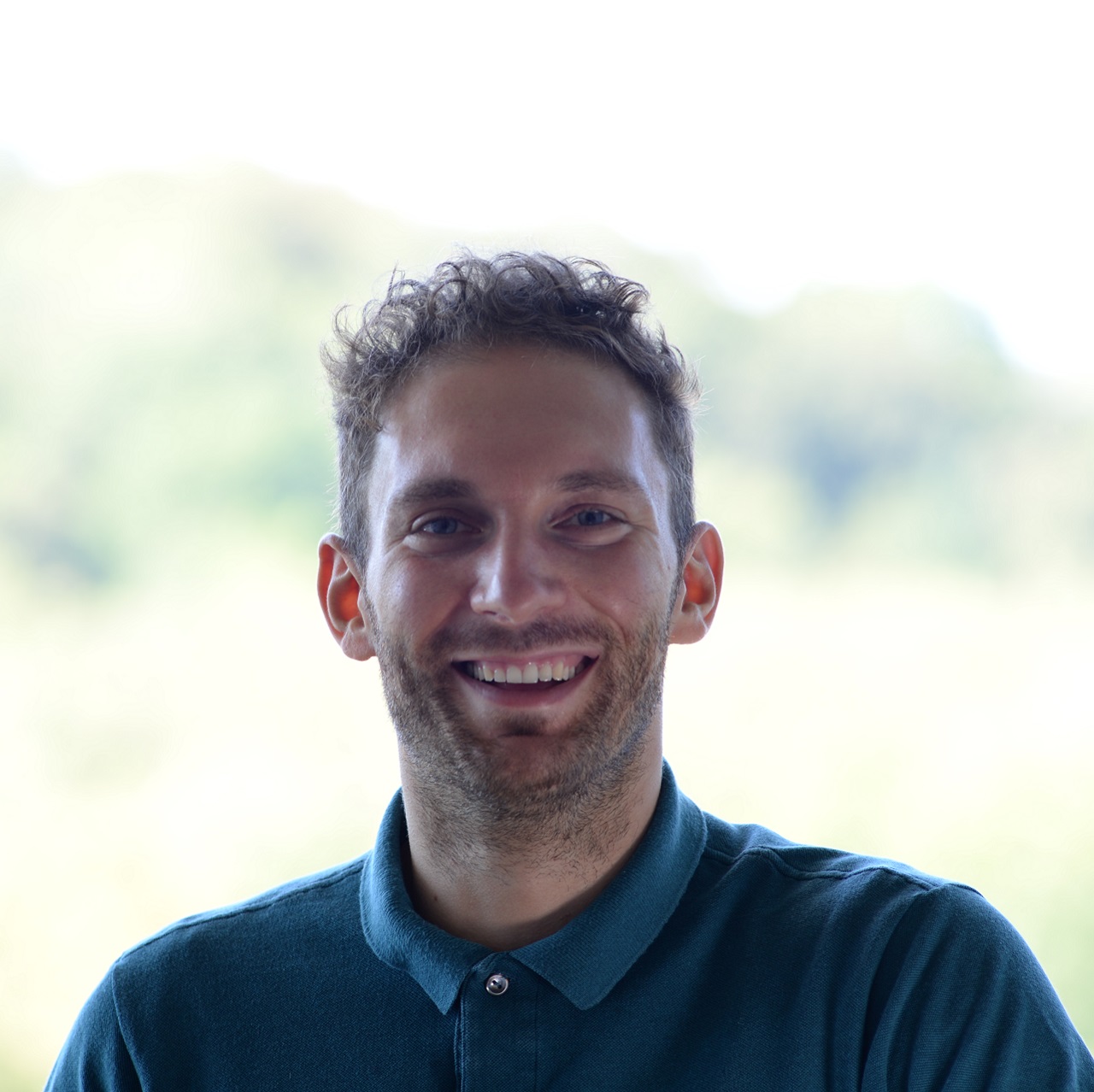}}]{Valerio Ortenzi}
Dr. Valerio Ortenzi qualified with a BSc in Automation and Automated Systems Engineering from Sapienza University of Rome in 2010. He went on to study for a MSc in Artificial Intelligence and Robotics Engineering, 2012, from Sapienza University of Rome, and then a PhD in Robotics in 2017, from the University of Birmingham. He then joined the Centre of Excellence in Robotic Vision at Queensland University of Technology, Brisbane, as a Research Fellow with Dist. Prof. Peter Corke. From October 2018 to April 2020, he was a PostDoctoral researcher at the University of Birmingham, UK. He is now a Research Scientist at the Max Planck Institute for Intelligence Systems, working with Director Katherine J. Kuchenbecker in the Haptic Intelligence Department.
His research interests focus on human-robot interaction and robotic manipulation. New directions of his work go into neuroscience and psychology of human grasping, and robotic vision.
\end{IEEEbiography}

\begin{IEEEbiography}[{\includegraphics[width=1in,height=1.25in,clip,keepaspectratio]{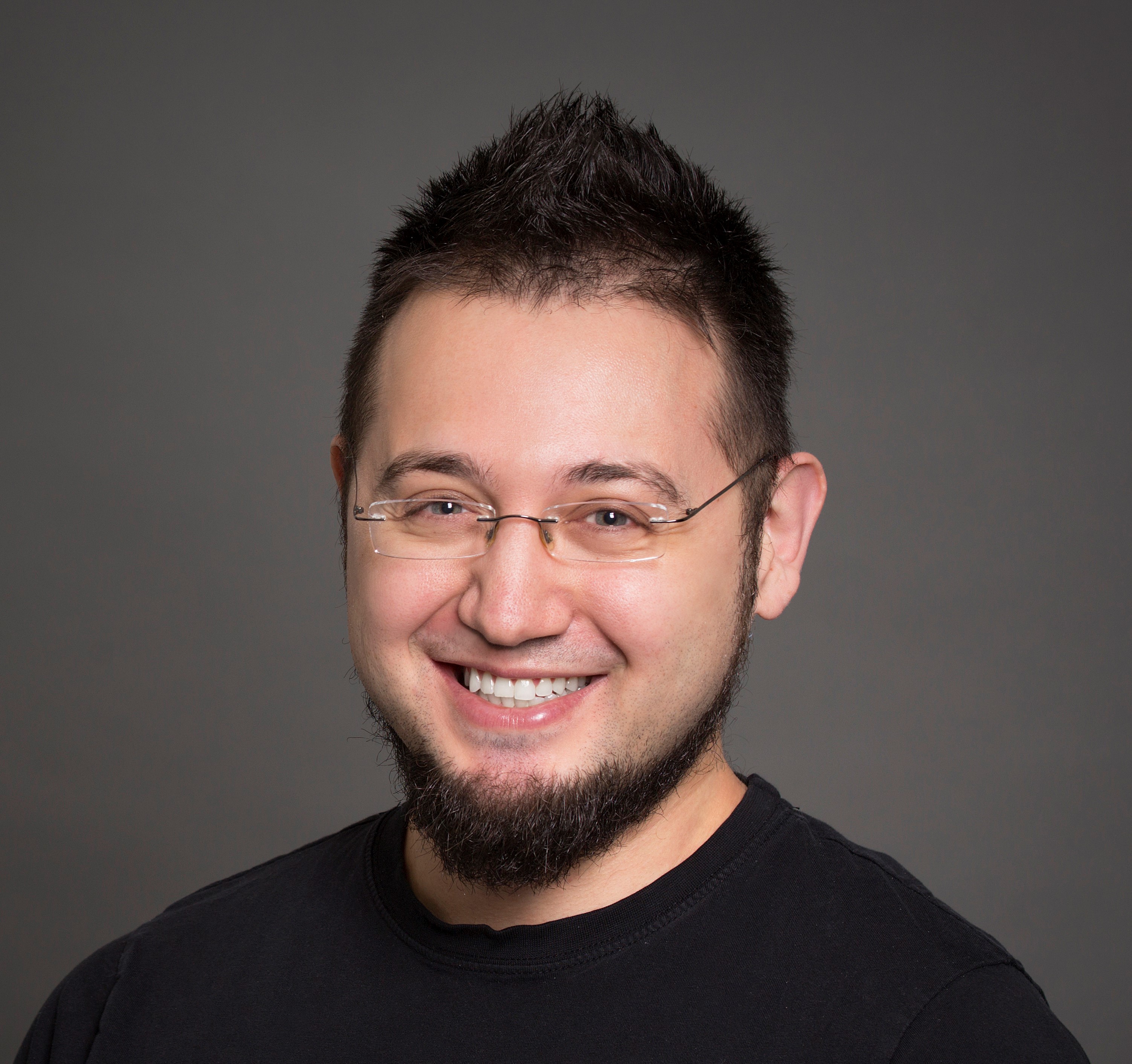}}]{Akansel Cosgun}
Dr. Akansel Cosgun conducts research in Robotics, Human-Robot Interaction (HRI), and Robot Learning. His research has been applied to mobile robots, robotic arms and self-driving cars with an emphasis on a systems view to problems. Akansel Cosgun received his Ph.D. in Robotics from Georgia Institute of Technology, USA in 2016. Since 2018, Dr. Cosgun has been a Research Fellow at Monash University, Australia. From 2019 to 2020 he was the Team Lead for Vision-based Manipulation for the Australian Centre for Robotic Vision (ACRV). Dr. Cosgun has previously worked at Honda Research, Toyota Infotechnology Center, Microsoft Research, and Savioke, a robotics start-up.
\end{IEEEbiography}
\vspace{-1.5cm}

\begin{IEEEbiography}[{\includegraphics[width=1in,height=1.25in,clip,keepaspectratio]{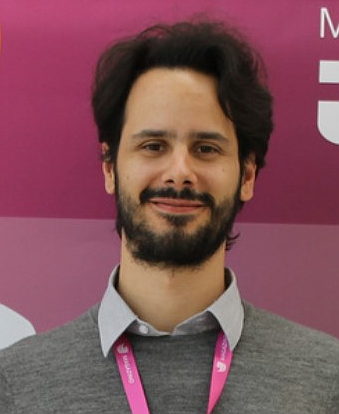}}]{Tommaso Pardi}
Tommaso Pardi qualified with a BSc in Computer engineering (2012) and a MSc in Autonomous and Control engineering (2015) from the University of Pisa.
After he graduated, he worked for nine months for a spin-off company of the University of Pisa as an Analyst Programmer. Then, he employed as a research fellow at the BioRobotics Institute of the School of Advanced Studies Sant'Anna in Pisa for one and a half years.
In his previous experiences, he worked on soft-robotics, grasping manipulation, robotic teleoperation, and UAV controls.
Since 2018, he has started his PhD at the Extreme Robotic Lab at the University of Birmingham to develop advance controllers for robotic cutting and resizing of nuclear wastes. In these years, he worked on grasp selection for post-grasp motions, motion planners for performing full-force tasks, and robot-human handover.
\end{IEEEbiography}
\vspace{-1.5cm}

\begin{IEEEbiography}[{\includegraphics[width=1in,height=1.25in,clip,keepaspectratio]{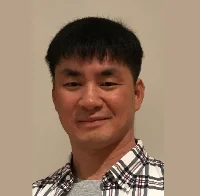}}]{Wesley P. Chan}
Dr. Wesley P. Chan is currently a Research Fellow at the Department of Electrical and Computer Systems Engineering at Monash University. His research interests include assistive robotics, human robot interaction, social robots, field robotics, and augmented/virtual reality. He received his B.A.Sc. in Engineering Physics in 2010 and M.A.Sc. in Biomedical Engineering in 2012 from the University of British Columbia (UBC). He then received his Ph.D. in Information Science and Technology in 2016 from the University of Tokyo. He conducted his M.A.Sc. and Ph.D. thesis research on a haptic-based robot object handover controller, and an affordance-focused framework for learning handover grasp configuration from observation and interaction respectively. He was awarded the Japanese Government MEXT scholarship and the Dean's Award for his doctoral research. He held postdoctoral researcher positions at the University of Tokyo and the UBC, prior to joining Monash University in 2019. In 2020, he was awarded the Foundation for Australia-Japan Studies grant to lead collaborative research with industry on socially conformant behaviours for autonomous robots.
\end{IEEEbiography}
\vspace{-1.5cm}

\begin{IEEEbiography}[{\includegraphics[width=1in,height=1.25in,clip,keepaspectratio]{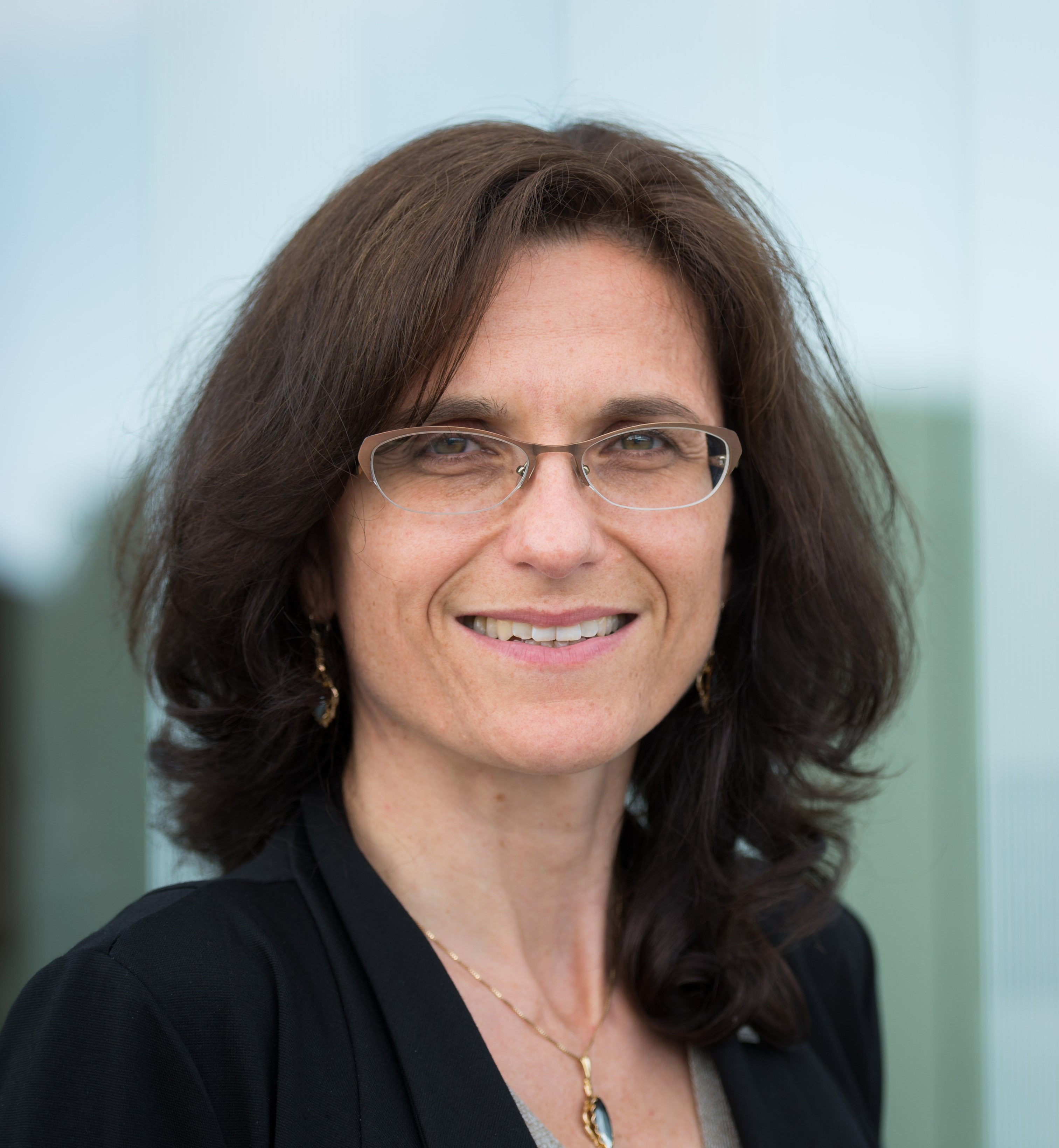}}]{Elizabeth Croft}
Professor Elizabeth A. Croft (B.A.Sc UBC ’88, M.A.Sc Waterloo ’92, Ph.D. Toronto ’95) is the Dean of Engineering at Monash University, in Melbourne, Australia.  Her research in industrial robotics and human-robot interaction advances the design of intelligent controllers and interaction methods that support safe and effective human-robot collaboration. Her recognitions include WXN’s top 100 most powerful women in Canada and the RA McLachlan Award for Professional Engineering. She is a Fellow of the Institute of Engineers Australia, the Canadian Academy of Engineers, Engineers Canada and the American Society of Mechanical Engineers.
\end{IEEEbiography}
\vspace{-1.5cm}

\begin{IEEEbiography}[{\includegraphics[width=1in,height=1.25in,clip,keepaspectratio]{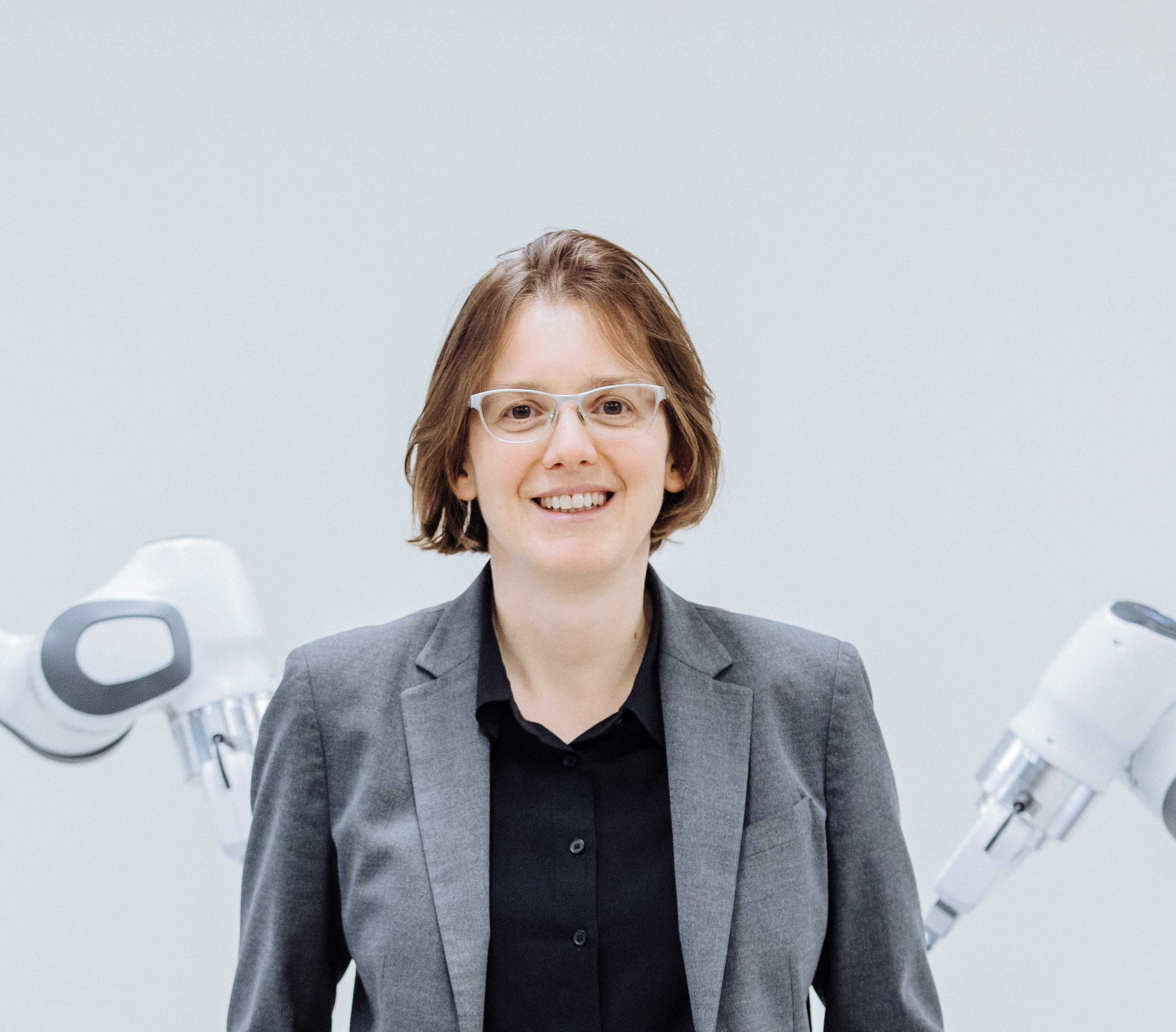}}]{Dana Kuli\'{c}}
Prof. Dana Kuli\'{c} conducts research in robotics and human-robot interaction (HRI), and develops autonomous systems that can operate in concert with humans, using natural and intuitive interaction strategies while learning from user feedback to improve and individualize operation over long-term use. Dana Kuli\'{c} received the combined B. A. Sc. and M. Eng. degree in electro-mechanical engineering, and the Ph. D. degree in mechanical engineering from the University of British Columbia, Canada, in 1998 and 2005, respectively. From 2006 to 2009, Dr. Kuli\'{c} was a JSPS Post-doctoral Fellow and a Project Assistant Professor at the Nakamura-Yamane Laboratory at the University of Tokyo, Japan. In 2009, Dr. Kuli\'{c} established the Adaptive System Laboratory at the University of Waterloo, Canada, conducting research in human robot interaction, human motion analysis for rehabilitation and humanoid robotics.  Since 2019, Dr. Kuli\'{c} is a professor and director of Monash Robotics at Monash University, Australia. In 2020, Dr. Kuli\'{c} was awarded the ARC Future Fellowship.  Her research interests include robot learning, humanoid robots, human-robot interaction and mechatronics.
\end{IEEEbiography}


\end{document}

%% file: table_final.tex
\begin{landscape}
\begin{table}
    \centering
    \centering
\begin{tabular}{? c ? c | c ? c | c | c | c ? c ? c | c | c ?c?c|c|c? c ?}
\hline
    
    \hline
    \rowcolor[HTML]{D0CECE}
         & \multicolumn{2}{c?}{Paradigm}  & \multicolumn{4}{c?}{Pre-Handover} & Handover & \multicolumn{3}{c?}{Physical Handover} & Post-HO & \multicolumn{3}{c?}{Metrics} & Test\\
        \rowcolor[HTML]{D0CECE}
         Paper & R2H & H2R & Comm. & Grasping & Motion & Perception &  Location & Perception & Grip Force & Error & Task & Task performance & \multicolumn{2}{c?}{User experience} &  Objects\\ 
         \rowcolor[HTML]{D0CECE}
         &  &  &  &  & &  &  &  &  & handling &  &  & Physiological & Subjective &  \\ \hline
         
         \rowcolor[HTML]{f0f0f5}
         \cite{Edsinger2007} & YES & YES & YES & YES & YES &   & Pre-planned &   &  &  & YES & {\begin{tabular}{c}failures,\\ grasp alignment error\end{tabular}} &  &  & 1 \\ \hline
         \cite{Choi2009} & YES &  & YES &  & YES & {\begin{tabular}{c}  laser pointer, camera,\\ laser range ﬁnder\end{tabular}} & Pre-planned & {\begin{tabular}{c}  force-sensing ﬁngers,\\ 6-axis force plate\end{tabular}} &  &  &  & {\begin{tabular}{c}detection time,\\ transport time,\\ placing/handing time,\\ grasping time,\\ total time, failures,\\ success rate\end{tabular}} &  & quest. & 3 \\ \hline
         \rowcolor[HTML]{f0f0f5}
         \cite{Huber2009} & YES &  &  &  & YES &   & Fixed & f/t sensors &  &  &  &  &  & quest. & 1 \\ \hline
         \cite{Sisbot2010} & YES &  &  &  & YES & camera & Pre-planned & force sensor &  &  &  &  &  &  & 1 \\ \hline
         \rowcolor[HTML]{f0f0f5}
         \cite{Fujita2010} & YES &  &  &  & YES &   & Fixed &   & &  &  &  & skin potential response & quest. & 1 \\ \hline
         \cite{Dehais2011} & YES &  &  &  & YES & {\begin{tabular}{c} sonars,\\ two laser range ﬁnders,\\ two stereo camera banks\end{tabular}}  & Pre-planned & {\begin{tabular}{c}contact sensors,\\ wrist force sensor\end{tabular}} &  &  &  &  & {\begin{tabular}{c}skin conductance,\\ deltoid muscle activity,\\ oculometry\end{tabular}} & quest. & 1 \\ \hline
         \rowcolor[HTML]{f0f0f5}
         \cite{Cakmak2011a}  & YES &  & YES &  & YES &   & Pre-planned &   &  &  &  &  &  & quest. & 5 \\ \hline
         \cite{Cakmak2011b}  & YES &  & YES &  & YES &   & Fixed &   &  &  &  &{\begin{tabular}{c} human waiting time, \\robot waiting time,\\ fluency\end{tabular}}  &  & quest. & 1 \\ \hline
         \rowcolor[HTML]{f0f0f5}
         \cite{Micelli2011} &  & YES &  & YES & YES & Kinect camera (RGBD) & Pre-planned & 6-axis f/t sensor &  &  &  & {\begin{tabular}{c}success rate,\\ completion time\end{tabular}} &  & quest. & 2 \\ \hline
         \cite{Bohren2011} & YES &  &  & YES & YES & tilting laser, stereo camera  & Fixed &   &  &  &  & completion time &  &  & 1 \\ \hline
         \rowcolor[HTML]{f0f0f5}
        \cite{Aleotti2012} & YES &  &  & YES & YES & {\begin{tabular}{c}laser scanner,\\   Kinect camera (RGBD)\end{tabular}} & Pre-planned &   &  &  &  & {\begin{tabular}{c}perception time,\\ completion time\end{tabular}} &  & quest. & 2 \\ \hline
        \cite{Mainprice2012} & YES &  &  &  & YES & camera, & Pre-planned &   &  &  &  & timing &  & quest. & 2 \\ \hline
        \rowcolor[HTML]{f0f0f5}
        \cite{Chan2013} & YES &  &  &  &  &   & Fixed & {\begin{tabular}{c}Pressure sensors,\\ 6-axis f/t sensor,\\ force sensing resistors\end{tabular}} & YES & &  & {\begin{tabular}{c}grip force, load force,\\ transfer time,\\ max load transfer rate\end{tabular}} &  & quest. & 1 \\ \hline
        \cite{Shi2013} & YES &  & YES &  & YES & {\begin{tabular}{c}3D range sensors,\\ cameras\end{tabular}} &  Online &   &  &  &  & success rate &  &  & 1 \\ \hline
        \rowcolor[HTML]{f0f0f5}
        \cite{Yamane2013} &  & YES &  &  & YES & motion capture &  Online &   &  &  &  &  &  &  & 1 \\ \hline
        \cite{Grigore2013} & YES &  & YES &  & YES & motion capture & Fixed &   &  &  &  & success rate &  &  & 1 \\ \hline
        \rowcolor[HTML]{f0f0f5}
        \cite{Aleotti2014} & YES &  & YES & YES & YES & {\begin{tabular}{c}laser scanner,\\   Kinect camera (RGBD)\end{tabular}} &  Pre-planned &   &  &  &  & success rate, timing &  & quest. & 3 \\ \hline
        \cite{Moon2014} & YES &  & YES &  &  &   & Fixed &   &  &  &  & timing &  & quest. & 1 \\ \hline
        \rowcolor[HTML]{f0f0f5}
        \cite{Koene2014} & YES & YES & YES & YES & YES & Kinect camera (RGBD) & Online & {\begin{tabular}{c}capacitive touch\\ sensors\end{tabular}} & YES &  &  & success rate &  & quest. & 1 \\ \hline
        \cite{Prada2014} & YES & YES & YES & YES & YES & Kinect camera (RGBD) & Online &   &  &  &  & success rate &  & quest. & 1 \\ \hline
        \rowcolor[HTML]{f0f0f5}
        \cite{Koene2014b} & YES & YES & YES & YES & YES & Kinect camera (RGBD) & Online &   &  &  & YES & {\begin{tabular}{c}Timings,\\ end position error\end{tabular}} &  & quest. & 1 \\ \hline
        \cite{Chan2014b} & YES &  &  & YES &  &   & Fixed & force sensor & YES &  &  & {\begin{tabular}{c}timing,\\ maximum excess\\ receiver load force\end{tabular}} &  &  & 1  \\ \hline
        \rowcolor[HTML]{f0f0f5}
        \cite{Chan2014} & YES &  &  & YES &  & motion capture & Fixed &   &  &  & YES &  &  &  & 3 \\ \hline
        \cite{Unhelkar2014} & YES &  & YES &  & YES & Laser scanner, RGBD & Fixed &   &  &  & YES & {\begin{tabular}{c}interaction time,\\ assistant idle time\end{tabular}} &  & quest. & 3* \\ \hline
        \rowcolor[HTML]{f0f0f5}
        \cite{Suay2015} & YES &  &  &  & YES & Kinect camera (RGBD) & Pre-planned &   &  &  &  &  & {\begin{tabular}{c}electromyogram\\ (EMG)\end{tabular}}  &  & 3 \\ \hline
\end{tabular}
\caption{Part I: Comparison of the papers presenting experiments with a real system. This table reports: the paradigm (robot-to-human R2H or human-to-robot H2R); the aspects of the pre-handover phase under investigation (communication, grasping, motion planning and control, perception); the sensors used; whether the handover location was fixed or fully pre-planned or adapted online; whether grip force was regulated and error handling dealt with during the physical handover; whether the experimental protocol foresaw a post-handover task; which metrics were utilised for the evaluation of the experiments (task metric, physiological, subjective); and how many objects were used in the handover tests. }
\label{table:partI}
\end{table}
\end{landscape}

\begin{landscape}
\begin{table}
    \centering
    \centering
\begin{tabular}{? c ? c | c ? c | c | c | c ? c ? c | c | c ?c?c|c|c? c ?}
\hline
    
    \hline
    \rowcolor[HTML]{D0CECE}
         & \multicolumn{2}{c?}{Paradigm}  & \multicolumn{4}{c?}{Pre-Handover} & Handover & \multicolumn{3}{c?}{Physical Handover} & Post-HO & \multicolumn{3}{c?}{Metrics} & Test\\
        \rowcolor[HTML]{D0CECE}
         Paper & R2H & H2R & Comm. & Grasping & Motion & Perception &  Location & Perception & Grip Force & Error & Task & Task performance & \multicolumn{2}{c?}{User experience} &  Objects\\ 
         \rowcolor[HTML]{D0CECE}
         &  &  &  &  & &  &  &  &  & handling &  &  & Physiological & Subjective &  \\ \hline
         
         \rowcolor[HTML]{f0f0f5}
         \cite{Admoni2014} & YES &  & YES &  & YES &   & Fixed &   & YES &  & YES & participant compliance &  & quest. & 1 \\ \hline
         \cite{Huang2015} & YES &  & YES &  &  & Kinect camera (RGBD) & Fixed & force sensor &  &  & YES & {\begin{tabular}{c}task completion time,\\ concurrent activity,\\ user idle time,\\ robot idle time\end{tabular}} &  & quest. & 1 \\ \hline
         \rowcolor[HTML]{f0f0f5}
         \cite{Parastegari2016} & YES &  &  &  &  & optical sensor & Fixed & Force sensor & YES & YES &  & forces & & quest. & 1 \\ \hline
         \cite{Maeda2016} & YES & YES &  &  & YES &   & Online &   &  &  & YES & success rate &  &  & 4 \\ \hline
         \rowcolor[HTML]{f0f0f5}
         \cite{Vahrenkamp2016} & YES &  &  & YES & YES &   & Online & Force sensor &  &  &  &  &  &  & 1 \\ \hline
         \cite{Kupcsik16} & YES &  &  &  & YES & Kinect camera (RGBD) & Fixed & Force sensor & YES &  &  &  &  & quest. & 1 \\ \hline
         \rowcolor[HTML]{f0f0f5}
         \cite{Medina2016} & YES &  &  &  & YES & motion capture & Online & force/torque sensor &  &  &  & forces, timing &  &  & 1 \\ \hline
         \cite{Gomez2017} & YES &  &  &  &  &   & Fixed & tactile sensors & YES & YES &  & forces &  &  & 3 \\ \hline
         \rowcolor[HTML]{f0f0f5}
         \cite{Bestick2017} & YES &  &  & YES & YES &   & Pre-planned &   &  &  & YES & estimated ergonomic cost & & quest. & 1 \\ \hline
         \cite{Peternel2017} & YES &  &  &  & YES & IMU suit &  Online & force/torque sensor &  &  &  &  & EMG &  & 1 \\ \hline
         \rowcolor[HTML]{f0f0f5}
         \cite{Konstantinova2017} & YES & YES &  &  &  &   & Fixed & force/torque sensor & YES &  &  & forces, jerk & &  & $>$6  \\ \hline
         \cite{Quispe2017} & YES &  & YES &  & YES & RGBD, rangefinder & Pre-planned &   &  &  &  &  &  & quest. & 1 \\ \hline
         \rowcolor[HTML]{f0f0f5}
         \cite{Meyer2017} & YES & YES & YES &  & YES & Camera &  Fixed & Force sensor & YES &  &  & {\begin{tabular}{c}movement speed,\\ reaction time,\\ manipulation time\end{tabular}} &  & quest. & 3 \\ \hline
         \cite{Controzzi2018} & YES &  &  &  &  &   & Fixed & force/torque sensor & YES &  &  & timing, acceleration, forces &  & quest. & 1 \\ \hline
         \rowcolor[HTML]{f0f0f5}
         \cite{Pan2018} &  & YES & YES & YES & YES & motion capture &  Online & force/torque sensor & YES & YES &  & forces, &  & quest. & 1 \\ \hline
         \cite{Bianchi2018} &  & YES &  & YES &  & IMU & Pre-planned &   &  &  &  & Success rate &  &  & 22 \\ \hline
         \rowcolor[HTML]{f0f0f5}
         \cite{Han2019} & YES &  & YES & YES & YES &   & Online &  {\begin{tabular}{c}force sensor,\\ force sensing resistor\end{tabular}} & YES &  & YES & completion time &  & quest. & 1  \\ \hline
         \cite{Wu2019} &  & YES &  &  & YES & motion capture & Online &   &  &  &  &  &  &  &  \\ \hline
         \rowcolor[HTML]{f0f0f5}
         \cite{Davari2019} & YES &  &  &  &  &   & Fixed & Tactile sensors &  & YES &  &  &  &  & 3 \\ \hline
         \cite{Sidiropoulos2019} & YES &  &  &  & YES & Kinect camera (RGBD) & Online &   & YES &  &  &  &  &  & 1 \\ \hline
         \rowcolor[HTML]{f0f0f5}
         \cite{Yang2020} &  & YES &  & YES & YES & RGBD &  Pre-planned &   &  &  &  &  {\begin{tabular}{c}success rate,\\ detection rate,\\ timing\end{tabular}}  &  & quest. & 1 \\ \hline
        \cite{Rosenberger2020} &  & YES &  & YES & YES & RGBD & Pre-planned &   &  &  &  & success rate &  &  & 13 \\ \hline
        \rowcolor[HTML]{f0f0f5}
        \cite{Ortenzi2020} & YES &  &  & YES &  &   & Fixed & force/torque sensor &  &  & YES & {\begin{tabular}{c}timing, post-handover\\ readjustments\end{tabular}} &  & quest. & 5\\ \hline
        \cite{Sanchez-Matilla2020} &  & YES &  & YES & YES & Intel RealSense (RGBD) & Online &  &  &  & YES & {\begin{tabular}{c}timings,\\ spatial accuracy\end{tabular}} &  &  & 4 \\ \hline
        \rowcolor[HTML]{f0f0f5}
        \cite{Ardon2020} & YES &  &  & YES & YES & Camera & Pre-planned &   &  &  & YES & accuracy &  &  & 10 \\ \hline
        \cite{Yang2020reactive} &  & YES &  & YES & YES & RGBD & Pre-planned &   &  &  &  & {\begin{tabular}{c}completion time,\\ grasp attempts,\\  success rate,\\ approach time\end{tabular}} &  & {\begin{tabular}{c}quest.,\\ open questions\end{tabular}} & 26 \\ \hline
         
\end{tabular}
\caption{Part II: Comparison of the papers presenting  experiments with a real system. This table reports: the paradigm (robot-to-human R2H or human-to-robot H2R); the aspects of the pre-handover phase under investigation (communication, grasping, motion planning and control, perception); the sensors used; whether the handover location was fixed or fully pre-planned or adapted online; whether grip force was regulated and error handling dealt with during the physical handover; whether the experimental protocol foresaw a post-handover task; which metrics were utilised for the evaluation of the experiments (task metric, physiological, subjective); and how many objects were used in the handover tests. }
\label{table:partII}
\end{table}
\end{landscape}